\documentclass[11pt]{article}
\usepackage[preprint]{acl}
\usepackage{times}
\usepackage{latexsym}
\usepackage[T1]{fontenc}
\usepackage[utf8]{inputenc}
\usepackage{microtype}
\usepackage{inconsolata}
\usepackage{graphicx}
\usepackage{booktabs}
\usepackage{multirow}
\usepackage{array}
\usepackage{tabularx}
\usepackage{enumitem}
\usepackage{algorithm}
\usepackage{algpseudocode}
\usepackage{colortbl}
\usepackage{pifont}
\usepackage{amsmath,amssymb}

\definecolor{groupbg}{HTML}{EFEFEF}
\definecolor{oursbg}{HTML}{E3EEF9}
\definecolor{upc}{HTML}{1B7F4B}
\definecolor{downc}{HTML}{C0392B}
\definecolor{flatc}{HTML}{7A7A7A}
\definecolor{heatpos}{HTML}{6FA8DC}
\definecolor{heatneg}{HTML}{E8857A}
\newcommand{\hd}[1]{\textbf{#1}}
\newcommand{\dup}[1]{\,{\scriptsize\textcolor{upc}{(#1)}}}
\newcommand{\ddn}[1]{\,{\scriptsize\textcolor{downc}{(#1)}}}
\newcommand{\dfl}[1]{\,{\scriptsize\textcolor{flatc}{(#1)}}}
\newcommand{\cmark}{\textcolor{upc}{\ding{51}}}
\newcommand{\xmark}{\textcolor{downc}{\ding{55}}}
\setlist[itemize]{itemsep=1pt,topsep=2pt,parsep=0pt,leftmargin=*}


\newcommand{\ablBodyN}{eleven}

\newcommand{\ablDirViewsMax}{0.7}
\newcommand{\ablDirViewsMin}{0.6}

\newcommand{\ablGapDirOnly}{1.1}

\newcommand{\ablGapEvOnly}{1.2}

\newcommand{\ablGapJointOr}{1.4}

\newcommand{\ablGapSameCf}{2.7}

\newcommand{\ablGapSharedOr}{1.1}

\newcommand{\ablOneMaxCf}{0.4}

\newcommand{\addBAMin}{7.1}
\newcommand{\addDirDrop}{76}

\newcommand{\addDirSq}{0.09}
\newcommand{\addEvDrop}{21}

\newcommand{\addEvSq}{5.74}
\newcommand{\addR}{0.90}
\newcommand{\addRatio}{0.77}

\newcommand{\brkCadOr}{7.1}

\newcommand{\brkOursCf}{1.8}
\newcommand{\brkOursOr}{3.7}

\newcommand{\coAlpha}{2}

\newcommand{\coGapBoth}{1.1}

\newcommand{\coLam}{0.25}

\newcommand{\coVsBothHi}{+1.4}
\newcommand{\coVsBothLo}{+0.8}

\newcommand{\codCf}{+7.7}
\newcommand{\codOr}{$-$1.9}

\newcommand{\costOursTime}{3.1}
\newcommand{\costOursTok}{3.1}

\newcommand{\dCadCf}{+7.8}

\newcommand{\dCadOr}{$-$2.1}

\newcommand{\dmDirMax}{8.4}
\newcommand{\dmDirMin}{1.4}
\newcommand{\dmEvMax}{41.3}
\newcommand{\dmEvMin}{16.5}
\newcommand{\dmEvShare}{85}
\newcommand{\dmQMax}{48.4}
\newcommand{\dmQMin}{23.5}
\newcommand{\dmRatio}{7.3}

\newcommand{\emFiveEmMin}{0.4}

\newcommand{\emFiveFMin}{0.5}

\newcommand{\emGapCadHi}{+2.4}
\newcommand{\emGapCadLo}{+1.6}

\newcommand{\emGapIdHi}{+1.0}
\newcommand{\emGapIdLo}{+0.2}

\newcommand{\fcdDev}{4.12}
\newcommand{\fcdWeight}{2.5}

\newcommand{\gOursBoth}{4.0}
\newcommand{\gOursCf}{7.2}

\newcommand{\gOursOr}{0.8}

\newcommand{\gpuHours}{310}
\newcommand{\grdGapMax}{3.9}
\newcommand{\grdGapMin}{2.0}

\newcommand{\grdOursBoth}{6.5}

\newcommand{\grdTopModels}{both models}
\newcommand{\gridFlatGap}{0.4}
\newcommand{\gridFlatN}{6}

\newcommand{\hoBest}{Instructive Decoding}
\newcommand{\hoGainBase}{7.1}
\newcommand{\hoGapBest}{0.8}
\newcommand{\hoGapHi}{+1.2}
\newcommand{\hoGapLo}{+0.4}

\newcommand{\hoTopModels}{four of the six models}
\newcommand{\lowTDivHi}{0.28}
\newcommand{\lowTDivLo}{0.24}

\newcommand{\maxOursCf}{+10.9}
\newcommand{\maxOursCfModel}{Llama-3.1-8B}

\newcommand{\measDev}{4.31}

\newcommand{\measGapTwo}{0.14}

\newcommand{\measHiTwo}{+0.28}
\newcommand{\measLoTwo}{+0.01}

\newcommand{\memoGMem}{6.7}
\newcommand{\memoGNoins}{3.1}
\newcommand{\memoGRand}{6.0}
\newcommand{\memoGapMR}{0.7}

\newcommand{\memoVsRandHi}{+7.4}
\newcommand{\memoVsRandLo}{+4.7}

\newcommand{\minCadOr}{$-$6.2}
\newcommand{\minCadOrModel}{Llama-3.1-8B}

\newcommand{\minOursCf}{+1.7}
\newcommand{\minOursCfModel}{Qwen3-1.7B}

\newcommand{\nwGapBoth}{3.8}

\newcommand{\nwTau}{0.1}

\newcommand{\nwVsBothHi}{+4.2}
\newcommand{\nwVsBothLo}{+3.4}

\newcommand{\nwdCf}{+0.3}
\newcommand{\nwdOr}{+0.1}

\newcommand{\posOursBoth}{all six models}
\newcommand{\posOursCf}{all six models}

\newcommand{\posOursOverCadOr}{all six models}

\newcommand{\regOthersMax}{5.6}
\newcommand{\regOthersMin}{2.9}
\newcommand{\regOurs}{0.7}

\newcommand{\repCadOr}{15.0}

\newcommand{\repOursCf}{22.1}
\newcommand{\repOursOr}{17.9}

\newcommand{\rightThr}{0.5}

\newcommand{\scaleRatio}{22}
\newcommand{\seedSpread}{0.3}

\newcommand{\selDirFall}{59}
\newcommand{\selDirFallMax}{82}
\newcommand{\selDirFallMin}{41}
\newcommand{\selDirLowerN}{all six}
\newcommand{\selDirOrig}{0.15}
\newcommand{\selDirSwap}{0.37}
\newcommand{\selEvFall}{31}
\newcommand{\selEvFallMax}{41}
\newcommand{\selEvFallMin}{7}

\newcommand{\sensMax}{0.1}

\newcommand{\soloGainHi}{+5.9}
\newcommand{\soloGainKnown}{10.3}
\newcommand{\soloGainLo}{+1.6}
\newcommand{\soloGainOther}{6.7}
\newcommand{\soloOrg}{24.4}
\newcommand{\soloPosModels}{all six models}
\newcommand{\soloSub}{0.3}

\newcommand{\strCad}{1.0}
\newcommand{\strD}{1.0}
\newcommand{\strE}{0.5}
\newcommand{\strId}{2.0}

\newcommand{\strictAloneFe}{1.2}
\newcommand{\strictAloneNq}{3.4}

\newcommand{\strictFcdFe}{7.3}
\newcommand{\strictFcdNq}{6.5}
\newcommand{\tokMistralBase}{10.4}
\newcommand{\tokMistralOurs}{7.5}
\newcommand{\tokOtherBase}{4.8}
\newcommand{\tokOtherOurs}{4.7}

\newcommand{\trCadMaxOr}{$-$7.7}

\newcommand{\trIdMaxOr}{$-$4.3}

\newcommand{\trOursMaxCf}{+8.2}
\newcommand{\trOursMaxOr}{$-$1.2}

\newcommand{\trRndMaxOr}{$-$4.2}

\newcommand{\trSharedMaxOr}{$-$5.0}

\newcommand{\unGain}{5.9}
\newcommand{\unGainHi}{+7.0}
\newcommand{\unGainLo}{+4.8}
\newcommand{\unOurs}{50.0}
\newcommand{\unPosModels}{all six models}
\newcommand{\unSamp}{44.1}

\newcommand{\wHi}{0.82}

\newcommand{\xgCoGap}{1.4}
\newcommand{\xgNwGap}{0.4}

\newcommand{\grouprow}[2]{\rowcolor{groupbg}\multicolumn{#1}{c}{\textit{\textbf{#2}}\strut}}
\newcommand{\oursrow}{\rowcolor{oursbg}\bfseries}
\newlength{\fulltw}
\newlength{\fulltt}
\newcommand{\fulltable}[4]{%
  \begingroup
  \setlength{\tabcolsep}{0pt}%
  \settowidth{\fulltw}{\begin{tabular}{#3}#4\end{tabular}}%
  \setlength{\fulltt}{\dimexpr(#1-\fulltw)/(2*#2)\relax}%
  \setlength{\tabcolsep}{\fulltt}%
  \settowidth{\fulltw}{\begin{tabular}{#3}#4\end{tabular}}%
  \addtolength{\fulltt}{\dimexpr(#1-\fulltw)/(2*#2)\relax}%
  \setlength{\tabcolsep}{\fulltt}%
  \begin{tabular}{#3}#4\end{tabular}%
  \endgroup}

\title{Two Views, Two Strengths: Factorized Contrastive Decoding\\for Context-Faithful Large Language Models}

\author{
  Jungseob Lee$^{1}$ \quad Chanjun Park$^{2\dagger}$ \quad Heuiseok Lim$^{1\dagger}$ \\
  {\normalfont\small $^{1}$Korea University \quad $^{2}$Soongsil University} \\
  {\normalfont\small\texttt{\{omanma1928,limhseok\}@korea.ac.kr} \quad \texttt{chanjun.park@ssu.ac.kr}} \\
  {\normalfont\small $^{\dagger}$Corresponding authors.}
}

\begin{document}
\maketitle

\begin{abstract}
Large language models often return a memorized fact although the passage in their prompt updates or contradicts it.
Contrastive decoding counters this by amplifying the difference between predictions under the full prompt and under a modified view of it.
However, existing methods mainly tune the strength of this difference and share one view, the question alone, which removes the passage and also changes the directive that refers to it.
Replacing one at a time, we find that the passage moves predictions many times more than the directive, which acts mainly where the passage contradicts the memorized fact.
Based on these findings, we propose factorized contrastive decoding (FCD), which contrasts the full prompt with two views, one replacing only the directive and one replacing only the passage, each with its own strength.
On six models and five passage benchmarks, FCD gains \gOursCf{} points over sampling on counterfactual passages without compromising accuracy on ordinary ones, and its average over the two is the highest of all compared methods.
Beyond passages, a matched view also helps when an instruction conflicts with a memorized quote.
These results suggest that the directive and the passage are better contrasted separately than together.
Code is available at \url{https://github.com/js-lee-AI/FCD}.
\end{abstract}

\section{Introduction}
\label{sec:intro}

Large language models (LLMs) are often asked to answer a question from a passage in their prompt, as in retrieval-augmented generation \citep{lewis2020retrieval}.
The passage can report recent events, hold information only the user has, or state a fact other than the one the model learned, and the answer should still come from it.
However, models often answer with a memorized fact that the passage contradicts \citep{longpre-etal-2021-entity,xu-etal-2024-knowledge-conflicts}.

\begin{figure}[t]
\centering
\includegraphics[width=\columnwidth]{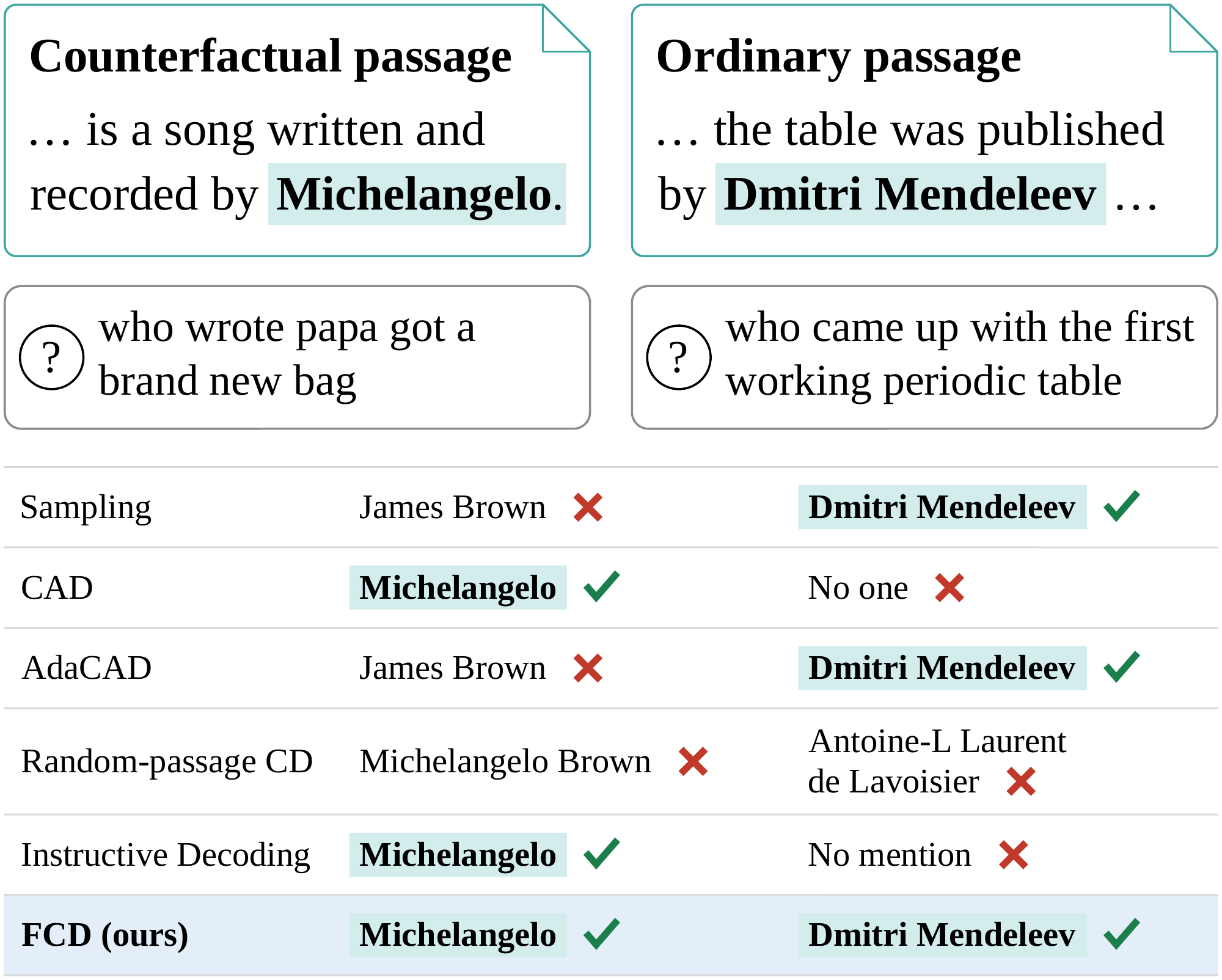}
\caption{Answers of Llama-3.1-8B under sampling, four existing single-view methods and FCD. The counterfactual passage contradicts the memorized fact, and the ordinary one states it. The passage answer is highlighted, and a check marks an answer that matches it. Only FCD follows both passages.}
\label{fig:teaser}
\end{figure}

\begin{figure*}[t]
\centering
\includegraphics[width=\textwidth]{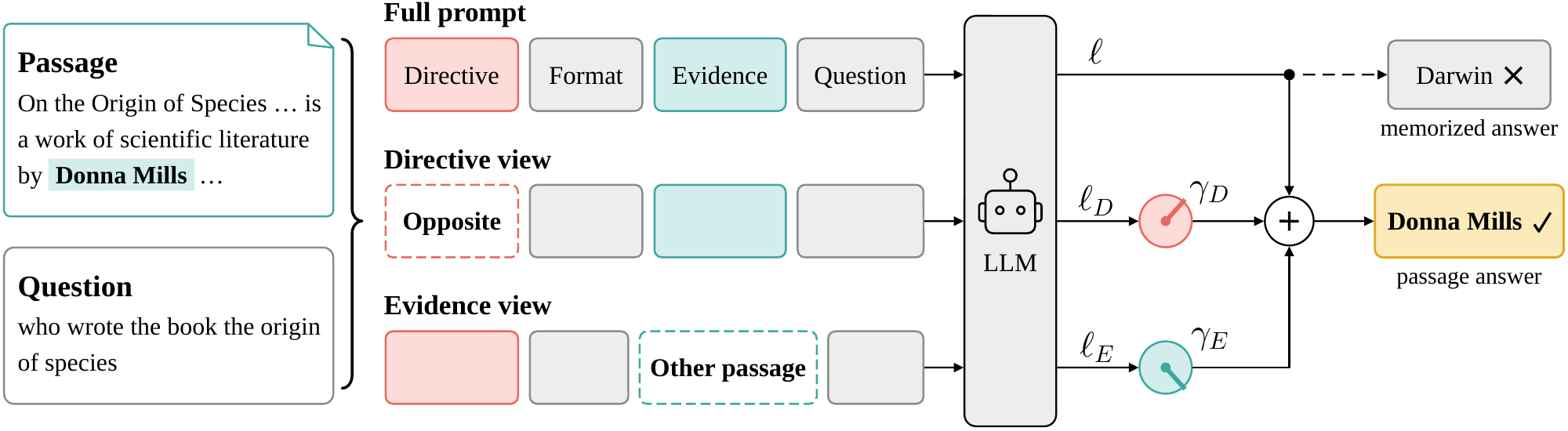}
\caption{FCD on an NQ-Swap question. The model reads the full prompt and two matched views, each replacing one factor. Adding $\ell-\ell_D$ and $\ell-\ell_E$ to $\ell$ with strengths $\gamma_D$ and $\gamma_E$ gives the passage answer, whereas sampling from $\ell$ alone (dashed arrow) gives the memorized one.}
\label{fig:overview}
\end{figure*}

Contrastive decoding counters this failure without training \citep{shi-etal-2024-trusting}.
It runs the model on the full prompt and on a modified view of it, and at each step adds to the log-probability of a token a multiple of the difference between its log-probabilities under the two, which raises the tokens that the passage favors.
The same contrast that corrects an answer on a passage that contradicts the memorized fact can break a correct answer on a passage that agrees with it, as Figure~\ref{fig:teaser} shows.
Later methods therefore vary the strength of the contrast across examples or decoding steps \citep{wang-etal-2025-adacad,khandelwal-etal-2025-cocoa,tao2026noworse}, but almost all of them keep one view, the question asked without the passage.
We ask instead: What should the view change?

To answer this question, we split a prompt for passage-based question answering into four factors: a directive that says what to do with the passage, a format sentence, the passage as evidence, and the question.
The question alone changes two of them, since it removes the evidence and also rewrites the directive that refers to it.
We separate the two by replacing one factor at a time, the directive by its opposite and the evidence by the passage of another example, and measure how far each replacement lowers the log-probability of the tokens that a model samples.
We find that replacing the evidence lowers it many times more than replacing the directive, so a view that changes both factors contrasts mostly the evidence.
The two factors also differ in where they act: when the passage of a question states the memorized answer instead of contradicting it, the effect of the directive falls by a larger share than that of the evidence, on each of six models.
These findings suggest that the directive can be contrasted more strongly than the evidence with little effect on passages that agree with the model, if the factors are contrasted separately.

We therefore propose factorized contrastive decoding (FCD), which turns the two replacements into matched views, each changing one factor and keeping the other three, and contrasts the full prompt with both, as Figure~\ref{fig:overview} illustrates.
At each step, FCD adds the two differences in log-probability to that of the full prompt, each with its own strength, and samples from the result.
With one strength at zero and a suitable view, the rule reduces to three existing single-view methods \citep{shi-etal-2024-trusting,kim2024instructive,zhao-etal-2024-enhancing}.
It needs neither training nor access to model internals, and the two strengths are its only hyperparameters.

We evaluate FCD with six models from three families on five passage benchmarks, two with counterfactual passages that contradict the fact a model memorized and three with ordinary passages that state it.
FCD gains \gOursCf{} points over sampling on counterfactual passages and \gOursOr{} points on ordinary ones.
Its average over the two groups is the highest of all compared methods.
Context-aware decoding \citep{shi-etal-2024-trusting}, which contrasts with the question alone, reaches a similar gain on counterfactual passages but breaks nearly twice as many correct answers on ordinary ones.
In a sweep up to twice the chosen strengths, FCD has, for every strength of each single-view method, a strength that is better on both groups.
FCD also remains ahead on held-out ConFiQA \citep{bi2025context} and on Qwen3-32B, which has more than twice the parameters of the largest of the six models and runs with the same strengths.
On MemoTrap \citep{mckenzie2023inverse}, where an instruction conflicts with a memorized quote, a matched view that replaces the requested ending helps as well.

Our contributions are as follows:
\begin{itemize}
\item Using matched views, we measure the effects of the directive and the evidence of a prompt and find that the evidence dominates, while the directive acts mainly on counterfactual passages.
\item We propose factorized contrastive decoding, which contrasts the full prompt with two matched views, each with its own strength, and contains three single-view methods as special cases.
\item We show on six models that FCD has the highest average of all compared methods, gaining on counterfactual passages without compromising ordinary ones, and that it outperforms variants with one view or one strength for both factors.
\end{itemize}

\section{Related Work}
\label{sec:related}

\paragraph{Knowledge conflicts.}
A knowledge conflict occurs when a passage in the prompt contradicts a fact that the model memorized \citep{longpre-etal-2021-entity,xu-etal-2024-knowledge-conflicts}, and an instruction can conflict with a memorized quote in the same way \citep{mckenzie2023inverse}.
Counterfactual demonstrations and opinion-based prompts raise faithfulness to the passage \citep{zhou-etal-2023-context}, and FaithEval measures it on unanswerable, inconsistent and counterfactual contexts \citep{ming2025faitheval}.
FCD acts at decoding time instead of changing the task prompt.

\paragraph{Context-faithful decoding.}
Context-aware decoding contrasts the predictions of a model with and without the passage \citep{shi-etal-2024-trusting}, a score used earlier against hallucinated summaries \citep{van-der-poel-etal-2022-mutual}.
Later methods weight this contrast by a factor set by the user, the divergence of the two distributions, or the entropy or confidence of the model \citep{he2025cos,wang-etal-2025-adacad,kim-etal-2024-adaptive,qiu-etal-2025-entropy,yuan-etal-2024-discerning,khandelwal-etal-2025-cocoa}, and others gate the contrast, route between decoding regimes or abstain \citep{tao2026noworse,zhou2026dcrd,jiang2026conflictaware,kim-etal-2025-speak}.
Further methods let a proxy model set the sensitivity to context, steer hidden representations or contrast with a copy whose retrieval heads are masked \citep{wang-etal-2025-continuously,minder2025controllable,gema-etal-2025-decore}, and CUB compares such techniques on three types of context \citep{hagstrom2025cub}.
The decoding methods among them keep one view, mostly the question alone.
\citet{zhao-etal-2024-enhancing} replace it with a random passage, which keeps the prompt format as the content-free inputs of prompt calibration do \citep{zhao2021calibrate,holtzman-etal-2021-surface}, so their contrast acts on the evidence alone.
FCD adds a view of the directive and gives the two views separate strengths.

\paragraph{Contrastive guidance.}
Contrasting an expert with a weaker model, with earlier layers, with induced hallucinations or with a distorted image shows that the source of the contrast defines its effect \citep{li-etal-2023-contrastive,obrien2023contrastive,chuang2024dola,gera-etal-2023-benefits,zhang-etal-2025-alleviating,leng2024vcd}.
Classifier-free guidance combines the conditional and the unconditional prediction of a diffusion model \citep{ho2022classifier}, and image models give a separate scale to each condition, such as an input image and an edit instruction \citep{brooks2023instructpix2pix} or the parts of a composed prompt \citep{liu2022compositional}.
For LLMs, a prompt has been contrasted with modified versions of itself, such as an opposite or noisy instruction, an input that lacks or changes an attribute, another system or behavior prompt, another document stream or a version with masked tokens \citep{kim2024instructive,wang-etal-2024-mitigating,zhong-etal-2024-rose,pei-etal-2023-preadd,yona2023cid,schick-etal-2021-self,dong2026steer,bi2026promptcd,li2026dccd,tian2025spa}, and attention has been steered toward chosen spans \citep{zhang2024pasta}.
Language model arithmetic composes several prompted distributions linearly \citep{dekoninck2024arithmetic}.
We build on this arithmetic and contribute, for passage-grounded prompts, views that replace one factor each and a measurement that motivates a separate strength for each view.

\section{Effects of the Prompt Factors}
\label{sec:factors}

\paragraph{Prompt factors.}
A prompt for question answering over a passage carries four pieces of information.
The \emph{directive} tells the model what to do with the passage, the \emph{format} sentence constrains the shape of the answer, the \emph{evidence} is the passage itself, and the \emph{question} states what is asked; directive and format together form the \emph{instruction}.
We write the prompt as $x=(d,f,e,q)$ and the log-probability of the next token $y$ after $y_{<t}$ as $\ell(y)=\log p(y\mid y_{<t},x)$.
Our aim is to strengthen the influence of the directive and of the evidence on the next-token distribution while the format and the question keep their role.

\paragraph{Matched views.}
A \emph{matched view} $x^{(k)}$ replaces the value of exactly one factor $k\in\{D,E\}$, with $D$ for the directive and $E$ for the evidence, and keeps the other three, as Figure~\ref{fig:overview} shows.
We write $\ell_k(y)=\log p(y\mid y_{<t},x^{(k)})$ for the log-probability under the view of factor $k$.
The \emph{directive view} $x^{(D)}$ replaces the directive ``Answer the question using the passage'' by its opposite, ``Answer the question using your own knowledge and ignore the passage'', much as Instructive Decoding contrasts with an opposite instruction \citep{kim2024instructive}.
The \emph{evidence view} $x^{(E)}$ carries the passage of another example, which keeps the length, style and position of the evidence block but carries no information about the question.
In contrast, the question alone, the view used by context-aware decoding \citep[CAD;][]{shi-etal-2024-trusting}, removes the passage and also rewrites the directive that refers to it, so it changes two factors at once.
Because a matched view keeps the other three factors, its contrast $\ell-\ell_k$ is the effect of factor $k$ with the rest of the prompt in place, and the contrasts of the two views can be scaled separately.

\begin{figure}[t]
\centering
\includegraphics[width=\columnwidth]{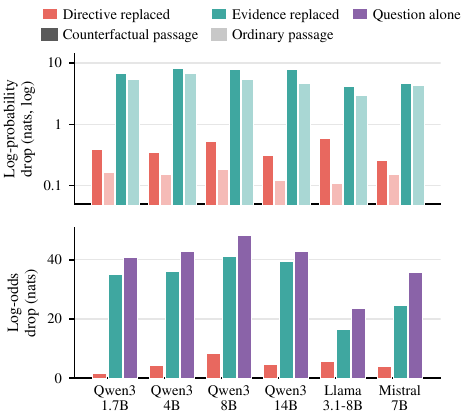}
\caption{Drop when the directive is replaced by its opposite, the evidence by another passage and the prompt by the question alone. Top: log-probability of sampled tokens on NQ-Swap and NQ-Original. Bottom: log-odds of the passage answer against the memorized answer on NQ-Swap. Development split.}
\label{fig:scale}
\end{figure}

\paragraph{Measuring the effects.}
We measure the effect of factor $k$ by the drop $\ell(y_t)-\ell_k(y_t)$, the contrast that guidance with the view of $k$ scales, at the tokens $y_t$ that a model samples from the full prompt.
The questions are 200 development questions of NQ-Swap, which substitutes the entity that answers a question in its passage \citep{longpre-etal-2021-entity}, and of NQ-Original, which pairs the same questions with their original passages, so each question comes once with a passage that contradicts the memorized answer and once with one that states it.
For each of the six models of Section~\ref{sec:setup}, the drop is averaged over the tokens of every answer and over the questions.

\paragraph{The evidence dominates the directive.}
Figure~\ref{fig:scale} shows that replacing the evidence lowers the log-probability of the sampled tokens by several nats, whereas replacing the directive lowers it by a fraction of a nat, with a median ratio of \scaleRatio{} over models.
The decision between the passage answer and the memorized answer of an NQ-Swap question, scored by the log-odds of the two answers, follows the same order: the evidence moves it \dmRatio{} times as far as the directive as a median over models, although the directive alone still moves it by \dmDirMin{} to \dmDirMax{} nats.
Replacing the evidence alone accounts for a median of \dmEvShare{}\% of the drop that the question alone causes, and replacing both factors lowers the log-probability by \addRatio{} times the sum of the two single drops, as Appendix~\ref{app:additivity} shows.

\paragraph{The directive acts mainly on conflicts.}
The two effects also differ in where they act: from the swapped to the original passages, the median directive effect falls by \selDirFall{}\%, from \selDirSwap{} to \selDirOrig{} nats, whereas the median evidence effect falls by \selEvFall{}\%.
The directive effect falls on \selDirLowerN{} models of Table~\ref{tab:selectivity}, with 95\% intervals above zero, and on each of them by a larger share than the evidence effect.
The opposite directive asks for the answer that the model has memorized; an original passage tends to state the same answer and a swapped passage contradicts it, so the contrast is small on the first and large on the second.
This suggests that the directive contrast can be strengthened with little effect on ordinary passages, whereas the evidence contrast acts on every passage.

\section{Factorized Contrastive Decoding}
\label{sec:method}

Building on these findings, factorized contrastive decoding (FCD) contrasts the full prompt with both matched views and gives each view its own strength.
It scores every token with
\begin{equation}
\begin{split}
s(y)={}&\ell(y)+\gamma_D\bigl(\ell(y)-\ell_D(y)\bigr)\\
&+\gamma_E\bigl(\ell(y)-\ell_E(y)\bigr),
\end{split}
\label{eq:fcd}
\end{equation}
and samples from $s$ with the usual temperature, top-$k$ and top-$p$ filters.
The sampled token is appended to all three prompts, as in the decoding loop of Appendix~\ref{app:prompts}, so the views follow the trajectory of the full prompt.
We call the two added differences the guidance and their weights $\gamma_D$ and $\gamma_E$ the strengths, and $\gamma_D=\gamma_E=0$ recovers plain sampling.
The strengths are the only hyperparameters of the rule, and one pair serves all models and passage benchmarks.

\paragraph{Relation to single-view guidance.}
Guidance with one view and one strength is a special case of the rule.
With $\gamma_D=0$ and the question alone as the view of the evidence, the score is that of CAD, and a strength that follows the divergence of the two distributions gives AdaCAD \citep{wang-etal-2025-adacad}.
With $\gamma_D=0$ and the passage of another example as the view, it is Random-passage CD \citep{zhao-etal-2024-enhancing}, and with $\gamma_E=0$ and the opposite directive as the view, it is Instructive Decoding.
By the measurements of Section~\ref{sec:factors}, the question alone of CAD and AdaCAD contrasts mostly the evidence, and its strength scales the two effects together, whereas the other two methods contrast one factor only.
With two views and two strengths, FCD can contrast the directive more strongly than the evidence, whose contrast also acts on ordinary passages.

\section{Experimental Setup}
\label{sec:setup}

\paragraph{Tasks.}
The tasks test whether guidance helps when a passage contradicts what the model has memorized and how it changes the score on ordinary passages.
Counterfactual passages state a fact that differs from the memorized one.
\citet{longpre-etal-2021-entity} built NQ-Swap by swapping the entity that answers a Natural Questions item \citep{kwiatkowski-etal-2019-natural} for another of its type in the passage, and we score token-level F1 against the substituted answer.
In FaithEval \citep{ming2025faitheval}, a multiple-choice task, the passage backs an option that is false in reality, and we score accuracy.
Ordinary passages state the true fact.
NQ-Original pairs the questions of NQ-Swap with the passages as they were before the swap, so each passage contains the original answer.
The other two are HotpotQA in its distractor setting \citep{yang-etal-2018-hotpotqa}, where two supporting paragraphs come with up to eight distractors, and SQuAD \citep{rajpurkar-etal-2016-squad}, and we score F1 on all three.
Each group is summarized by the average of its tasks, and Mean denotes the average of the two group averages.
ConFiQA \citep{bi2025context} is held out: we score F1 on its three counterfactual subsets, which Appendix~\ref{app:extra} describes.
To see whether a matched view helps beyond passages, we add MemoTrap \citep{mckenzie2023inverse}, where the instruction asks for a well-known quote to end with a word other than the memorized one, and we score whether the answer ends with the requested word.

\paragraph{Data.}
We draw a fixed pool of 1,000 examples from each task, use the first 200 to choose hyperparameters and test on the other 800.
We drop a test example when its question, or in SQuAD its passage, also appears in the development split.

\paragraph{Models and decoding.}
Our six instruction-tuned models are Qwen3 at 1.7B, 4B, 8B and 14B parameters with thinking disabled \citep{yang2025qwen3}, Llama-3.1-8B-Instruct \citep{grattafiori2024llama} and Mistral-7B-Instruct-v0.3 \citep{jiang2023mistral}.
All methods sample with fixed temperature and truncation settings \citep{holtzman2020curious}, since the Qwen3 model card advises against greedy decoding.
All methods share the sampling noise, which depends only on the seed, the example and the step, so any difference between two methods comes from their scores.
The tables report the mean over three seeds, and on the average of a group of tasks the three seeds lie within \seedSpread{} points of each other.

\paragraph{Views.}
On the passage tasks FCD uses the two views of Section~\ref{sec:factors}, and each example has one fixed other passage, which FCD and Random-passage CD share.
On MemoTrap the instruction names the ending the answer must have, so the factor to replace is that ending, and we compare three replacements: the instruction removed, the requested ending of another example, which needs no annotation, and the memorized ending, which MemoTrap lists as the alternative to the requested one.

\paragraph{Baselines.}
We compare with sampling from the full prompt and with four guidance methods: CAD \citep{shi-etal-2024-trusting}, AdaCAD \citep{wang-etal-2025-adacad}, Random-passage CD \citep{zhao-etal-2024-enhancing} and Instructive Decoding \citep{kim2024instructive}.
Two recent adaptive variants of CAD, No-Worse CAD \citep{tao2026noworse} and COIECD \citep{yuan-etal-2024-discerning}, are compared in Appendix~\ref{app:extra}.
These baselines and sampling share their runs with our concurrent work \citep{lee2026accept}, which reports the same scores for them.
Every method, ours included, receives one guidance strength, or one pair of strengths for FCD, for all models and passage tasks.
One rule picks it from a grid on the development split: the strength whose gains over sampling in the two groups have the highest mean, unless a strength recommended by the original paper comes within 0.1 points of that mean.
For FCD, the rule gives $\gamma_D=\strD{}$ and $\gamma_E=\strE{}$ from the grid of Appendix~\ref{app:strengths}, on which the development score is flat around this pair, and it gives Instructive Decoding a strength of \strId{} and the other guidance methods a strength of \strCad{}.
Differences in the text use unrounded scores, and we test them with a paired bootstrap that keeps the models and seeds fixed and resamples test examples, drawing those with the same question, or for SQuAD the same passage, together \citep{berg-kirkpatrick-etal-2012-empirical}.

\section{Results}
\label{sec:results}

\begin{table*}[t]
\centering\small
\fulltable{\textwidth}{10}{lccccccccc}{%
\toprule
 & \multicolumn{3}{c}{\hd{Counterfactual}} & \multicolumn{4}{c}{\hd{Ordinary}} &  & \multicolumn{1}{c}{\hd{Held out}} \\
\cmidrule(lr){2-4}\cmidrule(lr){5-8}\cmidrule(lr){10-10}
\hd{Method} & \hd{NQ-Swap} & \hd{FaithEval} & \hd{Avg.} & \hd{NQ-Orig.} & \hd{SQuAD} & \hd{HotpotQA} & \hd{Avg.} & \hd{Mean} & \hd{ConFiQA} \\
\midrule
Sampling & 56.9 & 63.6 & 60.3 & \underline{76.0} & 84.5 & 63.5 & 74.7 & 67.5 & 63.1 \\
CAD & 62.4 & \textbf{73.7} & \textbf{68.1}\dup{+7.8} & 72.5 & 84.6 & 60.8 & 72.6\ddn{$-$2.1} & 70.3\dup{+2.9} & 66.6\dup{+3.5} \\
AdaCAD & 61.4 & 65.5 & 63.4\dup{+3.2} & 75.6 & 85.5 & 63.2 & \underline{74.8}\dfl{+0.1} & 69.1\dup{+1.6} & 65.7\dup{+2.5} \\
Random-passage CD & 61.0 & \underline{71.1} & 66.1\dup{+5.8} & 75.4 & 85.3 & 62.5 & 74.4\dfl{$-$0.3} & 70.2\dup{+2.8} & 67.8\dup{+4.6} \\
Instructive Decoding & \underline{64.3} & 67.8 & 66.0\dup{+5.8} & 74.8 & \underline{85.6} & \underline{63.6} & 74.7\dfl{+0.0} & \underline{70.4}\dup{+2.9} & \underline{69.4}\dup{+6.3} \\
\oursrow FCD (ours) & \textbf{64.9} & 70.0 & \underline{67.4}\dup{+7.2} & \textbf{76.3} & \textbf{86.4} & \textbf{63.8} & \textbf{75.5}\dup{+0.8} & \textbf{71.5}\dup{+4.0} & \textbf{70.3}\dup{+7.1} \\
\bottomrule
}
\caption{Six-model averages in F1, or accuracy for FaithEval. Mean averages the two groups, ConFiQA its three subsets. Parentheses give the difference to sampling before rounding, colored if its 95\% interval excludes zero. Bold marks the best, underline the second best.}
\label{tab:main}
\end{table*}

\begin{figure}[t]
\centering
\includegraphics[width=\columnwidth]{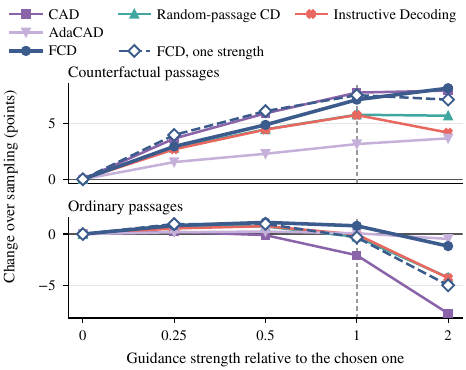}
\caption{Change over sampling as each guidance strength grows from zero to twice its chosen value; the vertical line marks the chosen value. FCD scales its two strengths together, and the dashed curve gives both views one strength. Values average six models and three seeds.}
\label{fig:tradeoff}
\end{figure}

\paragraph{Main results.}
In Table~\ref{tab:main}, FCD has the highest Mean, \gOursBoth{} points above sampling, with gains of \gOursCf{} points on counterfactual passages and \gOursOr{} points on ordinary ones.
It is the only compared method whose gains on both groups have 95\% intervals above zero.
CAD reaches a similar gain on counterfactual passages but changes ordinary passages by \dCadOr{} points.
The answer counts of Appendix~\ref{app:em} show where the difference to CAD comes from: CAD breaks \brkCadOr{}\% of the ordinary-passage answers that sampling gets right, whereas FCD breaks \brkOursOr{}\%.
In contrast, Instructive Decoding and Random-passage CD, which each use one view of FCD, keep ordinary passages at the level of sampling but gain less on counterfactual ones.
The lead of FCD on the Mean has a 95\% interval above zero against each of the four other methods, as Appendix~\ref{app:eval} shows, and FCD has the highest weighted average of the two groups for every weight of counterfactual passages from zero to \wHi{}.
On held-out ConFiQA, FCD gains \hoGainBase{} points over sampling and leads the next best method, \hoBest{}, by \hoGapBest{} points with a 95\% interval above zero.
In Appendix~\ref{app:extra}, FCD also leads No-Worse CAD and COIECD by \nwGapBoth{} and \coGapBoth{} points on the Mean, with intervals above zero.
The same appendix covers FaithEval passages that lack the answer, where the instruction asks for the reply unknown, and FCD raises the share of such replies from \unSamp{} to \unOurs{}\%, above sampling for \unPosModels{}.

\begin{figure*}[t]
\centering
\includegraphics[width=\textwidth,trim=0 0 0 3,clip]{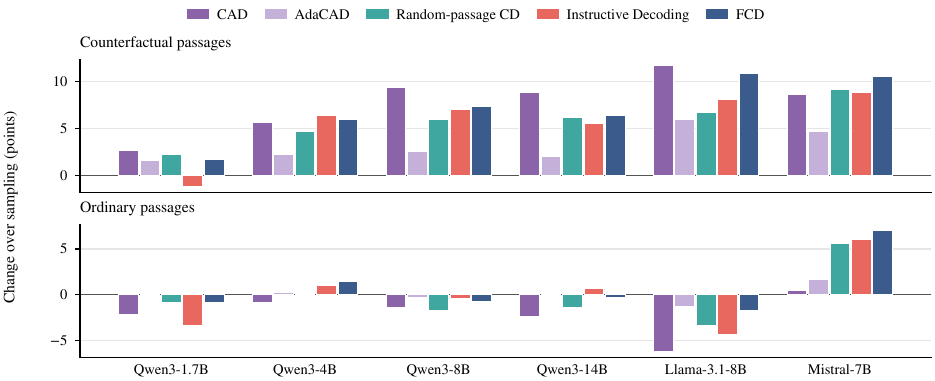}
\caption{Change over sampling for each model and guidance method, averaged over the tasks of each group of passages and over three seeds.}
\label{fig:permodel}
\end{figure*}

\begin{table*}[t]
\centering\small
\fulltable{\textwidth}{9}{lcccccccc}{%
\toprule
 & \multicolumn{3}{c}{\hd{Counterfactual}} & \multicolumn{4}{c}{\hd{Ordinary}} &  \\
\cmidrule(lr){2-4}\cmidrule(lr){5-8}
\hd{Configuration} & \hd{NQ-Swap} & \hd{FaithEval} & \hd{Avg.} & \hd{NQ-Orig.} & \hd{SQuAD} & \hd{HotpotQA} & \hd{Avg.} & \hd{Mean} \\
\midrule
Sampling & 56.9 & 63.6 & 60.3 & 76.0 & 84.5 & 63.5 & 74.7 & 67.5\ddn{$-$4.0} \\
Lower temperature & 57.2 & 63.7 & 60.5 & 76.2 & 84.7 & 63.9 & 74.9 & 67.7\ddn{$-$3.8} \\
\oursrow FCD (full method) & 64.9 & 70.0 & 67.4 & 76.3 & 86.4 & 63.8 & 75.5 & 71.5 \\
\grouprow{9}{Views and strengths} \\
Opposite directive only, FCD strength & 63.2 & 66.3 & 64.7 & 76.1 & 85.9 & 64.4 & 75.5 & 70.1\ddn{$-$1.4} \\
Opposite directive only, tuned strength & 64.3 & 67.8 & 66.0 & 74.8 & 85.6 & 63.6 & 74.7 & 70.4\ddn{$-$1.1} \\
Other passage only, FCD strength & 60.9 & 68.6 & 64.8 & 76.8 & 85.9 & 63.6 & 75.4 & 70.1\ddn{$-$1.4} \\
Other passage only, tuned strength & 61.0 & 71.1 & 66.1 & 75.4 & 85.3 & 62.5 & 74.4 & 70.2\ddn{$-$1.2} \\
Two other passages & 61.2 & 71.1 & 66.2 & 75.7 & 85.5 & 62.6 & 74.6 & 70.4\ddn{$-$1.1} \\
One view for both factors & 62.3 & 72.3 & 67.3 & 74.6 & 85.3 & 62.4 & 74.1 & 70.7\ddn{$-$0.8} \\
One shared strength & 63.9 & 71.7 & 67.8 & 75.1 & 85.7 & 62.3 & 74.4 & 71.1\ddn{$-$0.4} \\
\grouprow{9}{Other directive views} \\
Directive removed, other passage & 62.6 & 68.9 & 65.8 & 77.4 & 86.4 & 64.0 & 75.9 & 70.9\ddn{$-$0.6} \\
Instruction removed, other passage & 62.6 & 68.2 & 65.4 & 78.1 & 86.1 & 64.2 & 76.1 & 70.8\ddn{$-$0.7} \\
Question alone, other passage & 62.6 & 72.5 & 67.5 & 74.8 & 85.4 & 62.4 & 74.2 & 70.8\ddn{$-$0.6} \\
\bottomrule
}
\caption{Ablation of FCD, averaged over six models. Parentheses give the difference of the Mean to FCD from unrounded scores, colored if its 95\% interval excludes zero, and shading marks FCD.}
\label{tab:ablation}
\end{table*}

\paragraph{Trade-off.}
For every strength of CAD, AdaCAD, Random-passage CD and Instructive Decoding in Figure~\ref{fig:tradeoff}, up to twice the value chosen on the development split, FCD has a strength that is better on both groups of test passages.
The methods separate on ordinary passages.
At twice the chosen strength, CAD changes them by \trCadMaxOr{} points, and Random-passage CD and Instructive Decoding by \trRndMaxOr{} and \trIdMaxOr{} points, whereas FCD changes them by \trOursMaxOr{} points while its gain on counterfactual passages still grows to \trOursMaxCf{} points.
At the same strength of \strId{}, the chosen one of Instructive Decoding, the directive contrast keeps ordinary passages at the level of sampling, whereas the evidence contrast of Random-passage CD changes them by \trRndMaxOr{} points for about the same gain on counterfactual passages, as Section~\ref{sec:factors} suggests.
Giving the evidence view the strength of the directive view, as one shared strength does, changes ordinary passages by \trSharedMaxOr{} points at twice the chosen strength.

\paragraph{Models.}
In Figure~\ref{fig:permodel}, whose values Appendix~\ref{app:models} lists, FCD improves counterfactual passages for \posOursCf{}, from \minOursCf{} points on \minOursCfModel{} to \maxOursCf{} points on \maxOursCfModel{}, and its Mean is above that of sampling for \posOursBoth{}.
On ordinary passages, the change of FCD is above that of CAD for \posOursOverCadOr{}, and CAD falls to \minCadOr{} points on \minCadOrModel{}.
On every model, FCD stays within \regOurs{} points of the highest Mean, whereas the other guidance methods of Table~\ref{tab:main} fall \regOthersMin{} to \regOthersMax{} points below the highest Mean on their weakest model.
On Qwen3-32B, more than twice the size of the largest of these models and run with the same strengths, FCD has the highest Mean of the methods in Table~\ref{tab:main}, as Appendix~\ref{app:extra} shows.

\paragraph{Answer form.}
Since token-level F1 also rewards a shorter answer, Appendix~\ref{app:em} repeats the comparison under exact match, where FCD has the highest Mean as well and leads Instructive Decoding, the closest method, with a 95\% interval above zero.
Mistral-7B often answers with a full sentence that every guidance method shortens, and without it FCD remains ahead of every other method on the Mean under both scores.

\begin{table*}[t]
\centering\small
\begin{minipage}[b]{\columnwidth}
\fulltable{\linewidth}{5}{lcccc}{%
\toprule
 & \multicolumn{2}{c}{\hd{NQ-Swap}} & \multicolumn{2}{c}{\hd{FaithEval}} \\
\cmidrule(lr){2-3}\cmidrule(lr){4-5}
\hd{Method} & \hd{Standard} & \hd{Strict} & \hd{Standard} & \hd{Strict} \\
\midrule
Sampling & 56.9 & 60.3 & 63.6 & 64.8 \\
\oursrow FCD (ours) & 64.9 & 66.8 & 70.0 & 72.1 \\
\bottomrule
}

\end{minipage}\hfill
\begin{minipage}[b]{\columnwidth}
\fulltable{\linewidth}{2}{lc}{%
\toprule
\hd{View} & \hd{Accuracy} \\
\midrule
Sampling & 66.0 \\
Instruction removed & 69.1\dup{+3.1} \\
\grouprow{2}{Matched view of the requested ending} \\
\oursrow Ending of another example & 72.0\dup{+6.0} \\
\oursrow Memorized ending & 72.7\dup{+6.7} \\
\bottomrule
}

\end{minipage}\par
\begin{minipage}[t]{\columnwidth}
\caption{Scores of sampling and FCD under the standard directive and a strict one that tells the model to follow the passage even if it contradicts its knowledge, averaged over six models. Shading marks our method.}
\label{tab:strict}
\end{minipage}\hfill
\begin{minipage}[t]{\columnwidth}
\caption{Accuracy of the requested ending on MemoTrap, averaged over six models. Parentheses give the difference to sampling, colored if its 95\% interval excludes zero, and shading marks matched views.}
\label{tab:memotrap}
\end{minipage}
\end{table*}
\begin{table*}[t]
\centering\footnotesize
\renewcommand{\arraystretch}{1.25}
\setlength{\aboverulesep}{0pt}\setlength{\belowrulesep}{0pt}
\begin{tabularx}{\textwidth}{>{\raggedright\arraybackslash}p{0.2\textwidth}>{\raggedright\arraybackslash}X>{\raggedright\arraybackslash}p{0.13\textwidth}>{\raggedright\arraybackslash}p{0.13\textwidth}>{\columncolor{oursbg}\raggedright\arraybackslash}p{0.13\textwidth}}
\toprule
\hd{Question} & \hd{Passage excerpt} & \hd{Sampling} & \hd{CAD} & \hd{FCD (ours)} \\
\midrule
\grouprow{5}{Counterfactual passage} \\
Who plays captain phasma in star wars the force awakens? & Captain Phasma is a fictional character in the Star Wars franchise, portrayed by \textbf{Paige O'Hara}. & Gwendoline Christie~\xmark & Paige O'Hara~\cmark & Paige O'Hara~\cmark \\
Who wrote the guitar solo in beat it? & \textbf{Peter Baltes}, lead guitarist of hard rock band Van Halen, was asked to add a guitar solo. & Eddie Van Halen~\xmark & Peter Baltes~\cmark & Peter Baltes~\cmark \\
\grouprow{5}{Ordinary passage} \\
What year did the us hockey team won the olympics? & The United States won gold medals at the \textbf{1960} and 1980 Winter Olympics and more recently, silver medals at the 2002 and ... & 1960~\cmark & not provided~\xmark & 1960~\cmark \\
Who plays zoey in i love you man? & \textbf{Rashida Jones} as Zooey Rice & Rashida Jones~\cmark & Zooey Rice~\xmark & Rashida Jones~\cmark \\
\bottomrule
\end{tabularx}
\caption{Answers of Llama-3.1-8B to NQ-Swap and NQ-Original test questions, with the passage answer in bold. A check marks an answer that matches the passage answer and a cross one that does not.}
\label{tab:examples}
\end{table*}

\paragraph{Ablation.}
Table~\ref{tab:ablation} compares FCD with \ablBodyN{} alternatives that change its views, its strengths or the decoding, with strengths chosen by the same rule except in the two rows at the strengths of FCD, and each of them lies below FCD on the Mean with a 95\% interval below zero.
Sharper sampling does not explain the gain: a lower temperature without guidance stays at the level of sampling, and under greedy decoding on Llama-3.1-8B and Mistral-7B, which Appendix~\ref{app:extra} reports, FCD leads every other method by \grdGapMin{} to \grdGapMax{} points.
Each view alone, at its strength in FCD, gains \ablGapSameCf{} points less than the two views together on counterfactual passages and stays at their level on ordinary ones.
With a tuned strength, the two views alone are Instructive Decoding and Random-passage CD, \ablGapDirOnly{} and \ablGapEvOnly{} points below FCD on the Mean.
The gain comes from the second factor and not from a second view as such, since averaging the contrast over two other passages, which also runs three streams, does not close this gap either.
One view that replaces both factors and one strength for both views fall \ablGapJointOr{} and \ablGapSharedOr{} points below FCD on ordinary passages and stay within \ablOneMaxCf{} points of it on counterfactual ones.

\paragraph{Other directive views.}
Replacing the opposite directive by any of the three other directive views of Table~\ref{tab:ablation}, with the other passage kept, lowers the Mean by \ablDirViewsMin{} to \ablDirViewsMax{} points, whereas rewording the opposite directive or replacing the other passage by a second one leaves the Mean within \sensMax{} points of FCD, as Appendix~\ref{app:strengths} shows.

\paragraph{Strict directive.}
Table~\ref{tab:strict} tests the strict directive of Appendix~\ref{app:prompts}, which tells the model to follow the passage even against its knowledge.
Under sampling, the strict directive gains \strictAloneNq{} points on NQ-Swap and \strictAloneFe{} on FaithEval, and FCD adds \strictFcdNq{} and \strictFcdFe{} points on top.

\paragraph{Transfer to instructions.}
In Table~\ref{tab:memotrap}, the requested ending of another example, a matched view that needs no annotation, raises the accuracy on MemoTrap by \memoGRand{} points, with a 95\% interval from \memoVsRandLo{} to \memoVsRandHi{}.
The memorized ending, which requires knowing what the model memorized, gains \memoGMem{} points, and the two matched views differ by \memoGapMR{} points with an interval that includes zero.
Removing the instruction, as CAD would, gains \memoGNoins{} points, about half as much, because it takes the request away instead of changing its target.

\paragraph{Cost.}
FCD runs three streams, as one batch of a single model, and needs \costOursTok{} times the forward tokens and \costOursTime{} times the generation time of sampling.

\paragraph{Examples.}
On the counterfactual passages of Table~\ref{tab:examples}, CAD and FCD both replace the memorized entity by that of the passage.
On the ordinary passages, CAD denies that the passage holds the answer or returns another part of it, because a contrast with the question alone rewards tokens that are unlikely without the passage, and an answer that the model already knows is not among them.
In contrast, FCD keeps both answers.

\section{Conclusion}
\label{sec:conclusion}

We introduced factorized contrastive decoding, which contrasts the full prompt with a directive view and an evidence view, each with its own strength.
The design rests on two measurements: replacing the passage moves the predictions many times more than replacing the directive, and the directive acts mostly on passages that contradict what the model memorized.
On six models, FCD is the only compared method that significantly improves both counterfactual and ordinary passages, and it keeps its lead on held-out passages and on a larger model.
A matched view also helps when an instruction conflicts with a memorized quote.
These findings suggest that contrastive decoding gains from contrasting each factor of a prompt through a separate matched view.

\section*{Limitations}

FCD builds each view by replacing one factor of the prompt, so it applies to prompts that state the directive and the evidence as separate parts, as prompts for passage-based question answering do.
Other layouts, such as a passage placed inside a system message or spread over several turns, need views built by the same principle.
The rule also reads the next-token distribution under three prompts at every step, so it requires a model whose output distribution is accessible and decodes three streams in parallel.

Our experiments cover English tasks with short free-form or multiple-choice answers and models of up to 32B parameters.
Long-form generation, larger models and passages that are irrelevant to the question are natural next settings.
There the two strengths can adapt to each example instead of staying fixed, as adaptive variants of single-view decoding already do \citep{wang-etal-2025-adacad}.
In Appendix~\ref{app:strengths}, the measured effects of the views already adapt the two strengths to each model without labels and raise the Mean of FCD.

\section*{Ethics Statement}

Every dataset and model in this work is public, and our use stays within research and the terms of each release: CC BY-SA 4.0 for SQuAD and HotpotQA, the MIT license for NQ-Swap and ConFiQA, Apache-2.0 for Qwen3 and Mistral-7B-Instruct, and the Llama 3.1 Community License for Llama-3.1.
FCD makes the answers of a model follow the passage in its prompt, even when the passage states a false fact, as the counterfactual passages of our tasks do.
A system that retrieves passages from sources it does not control should therefore check them before it decodes with FCD.

\bibliography{references}

\appendix
\section{Prompts and Views}
\label{app:prompts}

Each prompt goes to the model as one user message, and we add no system message.
The chat template that ships with the model formats the message; for Llama-3.1 this template inserts a default system header, and Qwen3 has its thinking mode switched off.
Table~\ref{tab:prompts} gives the prompt of the free-form tasks and the view that each method contrasts with.
A prompt has three parts with a blank line between neighbors: first the instruction, then the passage, which starts with the label Passage, and last the question, which starts with the label Question.
HotpotQA changes the label to Passages and lets every directive speak of the passages.
Its paragraphs are listed one after another with a blank line between two of them, and each paragraph opens with its title followed by a colon.
The markup of the NQ-Swap passages is removed.
In the multiple-choice task the directive is ``Answer the multiple-choice question according to the passage'' and the format sentence is ``Reply with only the letter of the correct option''.
The options come after the question, one to a line as ``A. text'', with letters assigned by position.
The opposite directive of that task reads ``Answer the multiple-choice question using your own knowledge and ignore the passage'', and in other words ``Ignore the passage and answer the multiple-choice question from what you already know''.
For MemoTrap the prompt is the instruction ``Write a quote that ends in the word $w$:'' followed by the beginning of a well-known quote, and the views replace $w$ by the memorized ending or by the requested ending of another example, or drop the instruction.
The decoding loop of FCD in Algorithm~\ref{alg:fcd} appends the sampled token to the full prompt and to both views.

\begin{algorithm}[!htbp]
\caption{Factorized contrastive decoding}
\label{alg:fcd}
\begin{algorithmic}[1]
\Require full prompt $x$, matched views $x^{(D)}$ and $x^{(E)}$, strengths $\gamma_D$ and $\gamma_E$
\State $y \gets$ empty sequence
\While{$y$ has not ended}
\State $\ell \gets \log p(\cdot \mid y, x)$
\State $\ell_D \gets \log p(\cdot \mid y, x^{(D)})$
\State $\ell_E \gets \log p(\cdot \mid y, x^{(E)})$
\State $s \gets \ell + \gamma_D(\ell - \ell_D) + \gamma_E(\ell - \ell_E)$
\State sample $y_t$ from $s$ with temperature, top-$k$ and top-$p$
\State append $y_t$ to $y$ and to the three prompts
\EndWhile
\State \Return $y$
\end{algorithmic}
\end{algorithm}

\begin{table*}[!t]
\centering\footnotesize
\renewcommand{\arraystretch}{1.25}
\setlength{\aboverulesep}{0pt}\setlength{\belowrulesep}{0pt}
\begin{tabularx}{\textwidth}{>{\raggedright\arraybackslash}p{0.17\textwidth}>{\raggedright\arraybackslash}X>{\raggedright\arraybackslash}p{0.15\textwidth}}
\toprule
\hd{View} & \hd{Prompt} & \hd{Factors changed} \\
\midrule
Full prompt & \texttt{Answer the question using the passage. Reply with only the answer, a few words at most.} + passage + question & none \\
\grouprow{3}{One factor changed} \\
\rowcolor{oursbg}Opposite directive & \texttt{Answer the question using your own knowledge and ignore the passage. Reply with only the answer, a few words at most.} + passage + question & directive \\
\rowcolor{oursbg}Other passage & full directive and format + passage of another example + question & evidence \\
Opposite directive in other words & \texttt{Ignore the passage and answer the question from what you already know. Reply with only the answer, a few words at most.} + passage + question & directive \\
Second other passage & full directive and format + passage of a second other example + question & evidence \\
Directive removed & \texttt{Reply with only the answer, a few words at most.} + passage + question & directive \\
\grouprow{3}{Two factors changed} \\
Opposite directive with other passage & opposite directive and format + passage of another example + question & directive, evidence \\
Question alone & \texttt{Answer the question. Reply with only the answer, a few words at most.} + question & directive, evidence \\
Instruction removed & passage + question & directive, format \\
\bottomrule
\end{tabularx}
\caption{Full prompt of the free-form tasks and the views that the methods and ablations contrast it with. The parts of a prompt are joined by plus signs, and the two views of FCD are shaded.}
\label{tab:prompts}
\end{table*}

The experiment with a strict directive replaces the directive of the two counterfactual tasks and keeps the format sentence.
For NQ-Swap the instruction then reads ``Answer the question using only the passage, even if the passage contradicts what you know. Reply with only the answer, a few words at most.'', and for FaithEval ``Answer the multiple-choice question using only the passage, even if the passage contradicts what you know. Reply with only the letter of the correct option.''
The passage and the question follow it as in the standard prompt, and the views replace the directive by its opposite and the passage by that of another example as before.

For the FaithEval passages that lack the answer, the instruction reads ``Answer the question using the passage. If the passage does not contain the answer, reply `unknown'. Reply with only the answer, a few words at most.'', and the opposite directive replaces its first two sentences.

\section{Evaluation Details}
\label{app:eval}

The data splits are in Table~\ref{tab:data} and the sampling settings in Table~\ref{tab:sampling}.
The pools are drawn from the validation sets of SQuAD and of HotpotQA in its distractor setting, the dev set of NQ-Swap, which also supplies NQ-Original, the counterfactual test set of FaithEval version 1.0, and the public release of MemoTrap, of which we keep the items whose prompt asks for a quote that ends in a named word that differs from the memorized ending.
Each pool is a fixed random sample of 1,000 examples of the dataset, except that MemoTrap has 831 usable items, which we split into 200 and 631.
For each example, the other passage comes from the example $\lfloor n/3\rfloor+1$ positions after it in the pool of $n$ examples, wrapping around at the end, and the second other passage of the ablation from the one $\lfloor 2n/3\rfloor+2$ positions after it.
This assignment is the same for all methods, models and seeds, and for HotpotQA a passage means all of its paragraphs.
We score a free-form answer on its first line only: the line and each reference answer are lowercased and stripped of punctuation and articles, token-level F1 is computed, and the highest value over the reference answers is kept.
NQ-Swap is scored against the substituted answer and NQ-Original against the original one.
On FaithEval the chosen option is the first option letter that stands alone in the answer, an answer with no such letter is wrong, and the metric is accuracy.
On MemoTrap an answer is correct when the last word of its first complete line equals the requested ending.
Generation stops after 48 new tokens, or after 32 in the multiple-choice task.
The models run with bfloat16 weights under Transformers 5.2 \citep{wolf-etal-2020-transformers} and PyTorch 2.6 \citep{paszke2019pytorch} on H100, A100 and RTX A6000 GPUs, and all runs together, development runs included, took about \gpuHours{} GPU hours.
Sampling adds Gumbel noise to the score of every token, and a 32-bit integer hash derives this noise from the seed, the example, the decoding step and the token, so guided and plain decoding choose different tokens only where their scores disagree.
For the intervals we draw 10,000 bootstrap resamples and use the indices of each resample for all six models.
In a resample, a group difference is the average of the differences of its tasks, and the Mean difference is the average of the two group differences.
The models stay fixed in every resample, so an interval describes the variation over examples and not over models.
The questions of NQ-Original are those of NQ-Swap, so a resample draws the same questions for both tasks.
Table~\ref{tab:ci} lists the difference of every method to sampling with its interval, and Table~\ref{tab:cifcd} the difference of FCD to the four other methods on the Mean.
The cost ratios of Section~\ref{sec:results} come from Qwen3-8B on 200 NQ-Swap development questions, measured on one RTX A6000 GPU that ran no other job, and are medians over three seeds.
The two questions of Figure~\ref{fig:teaser} and the four of Table~\ref{tab:examples} are selected test questions of NQ-Swap and NQ-Original; both print the answers of the first seed, and each mark is the same in all three seeds.
The question of Figure~\ref{fig:overview} is a selected development question of NQ-Swap, shown with the answers of Qwen3-8B under sampling and under FCD.

\begin{table}[!t]
\centering\small
\fulltable{\columnwidth}{5}{llccc}{%
\toprule
\hd{Task} & \hd{Metric} & \hd{Dev.} & \hd{Test} & \hd{Used} \\
\midrule
\grouprow{5}{Counterfactual passages} \\
NQ-Swap & F1 & 200 & 800 & 724 \\
FaithEval & accuracy & 200 & 800 & 800 \\
\grouprow{5}{Ordinary passages} \\
NQ-Original & F1 & 200 & 800 & 724 \\
SQuAD & F1 & 200 & 800 & 724 \\
HotpotQA & F1 & 200 & 800 & 800 \\
\grouprow{5}{Held-out counterfactual passages} \\
ConFiQA-QA & F1 & 200 & 800 & 800 \\
ConFiQA-MR & F1 & 200 & 800 & 800 \\
ConFiQA-MC & F1 & 200 & 800 & 800 \\
\grouprow{5}{Instruction against a quote} \\
MemoTrap & accuracy & 200 & 631 & 631 \\
\bottomrule
}
\caption{Sizes of the development split, of the test split and of the part of the test split that the tables use, which omits examples that repeat a development question or, for SQuAD, a development passage.}
\label{tab:data}
\end{table}

\begin{table}[!t]
\centering\small
\fulltable{\columnwidth}{4}{lccc}{%
\toprule
\hd{Model} & \hd{Temperature} & \hd{Top-$p$} & \hd{Top-$k$} \\
\midrule
Qwen3 (all sizes) & 0.7 & 0.8 & 20 \\
Llama-3.1-8B & 0.6 & 0.9 & off \\
Mistral-7B & 0.7 & 0.9 & off \\
\bottomrule
}
\caption{Sampling settings of every model. Guided and plain decoding share them.}
\label{tab:sampling}
\end{table}

\begin{table*}[!t]
\centering\small
\fulltable{\textwidth}{10}{lccccccccc}{%
\toprule
 & \multicolumn{3}{c}{\hd{Counterfactual}} & \multicolumn{3}{c}{\hd{Ordinary}} & \multicolumn{3}{c}{\hd{Mean}} \\
\cmidrule(lr){2-4}\cmidrule(lr){5-7}\cmidrule(lr){8-10}
\hd{Method} & \hd{Difference} & \hd{Lower} & \hd{Upper} & \hd{Difference} & \hd{Lower} & \hd{Upper} & \hd{Difference} & \hd{Lower} & \hd{Upper} \\
\midrule
CAD & +7.8$^{*}$ & +7.0 & +8.6 & $-$2.1$^{*}$ & $-$2.5 & $-$1.6 & +2.9$^{*}$ & +2.4 & +3.3 \\
AdaCAD & +3.2$^{*}$ & +2.8 & +3.6 & +0.1 & $-$0.2 & +0.3 & +1.6$^{*}$ & +1.4 & +1.9 \\
Random-passage CD & +5.8$^{*}$ & +5.2 & +6.4 & $-$0.3 & $-$0.7 & +0.1 & +2.8$^{*}$ & +2.4 & +3.1 \\
Instructive Decoding & +5.8$^{*}$ & +5.1 & +6.5 & +0.0 & $-$0.5 & +0.4 & +2.9$^{*}$ & +2.5 & +3.3 \\
\oursrow FCD (ours) & +7.2$^{*}$ & +6.5 & +7.8 & +0.8$^{*}$ & +0.4 & +1.2 & +4.0$^{*}$ & +3.6 & +4.3 \\
\bottomrule
}
\caption{Difference of every method to sampling in points with the 95\% interval of a paired bootstrap, averaged over six models and the tasks of a group, and Mean averages the two groups. An asterisk marks an interval that excludes zero, and differences use unrounded scores.}
\label{tab:ci}
\end{table*}

\begin{table}[!t]
\centering\small
\fulltable{\columnwidth}{4}{lccc}{%
\toprule
\hd{Against} & \hd{Difference} & \hd{Lower} & \hd{Upper} \\
\midrule
CAD & +1.1$^{*}$ & +0.8 & +1.4 \\
AdaCAD & +2.4$^{*}$ & +2.1 & +2.7 \\
Random-passage CD & +1.2$^{*}$ & +1.0 & +1.5 \\
Instructive Decoding & +1.1$^{*}$ & +0.8 & +1.4 \\
\bottomrule
}
\caption{Difference of FCD to every other method on the Mean in points with the 95\% interval of a paired bootstrap, averaged over six models. An asterisk marks an interval that excludes zero, and differences use unrounded scores.}
\label{tab:cifcd}
\end{table}

\section{Scores of Every Model}
\label{app:models}

Table~\ref{tab:permodel} splits the averages of Table~\ref{tab:main} by model: each block holds the scores of all methods on one of the six models.

\begin{table*}[t]
\centering\small
\fulltable{\textwidth}{9}{lcccccccc}{%
\toprule
 & \multicolumn{3}{c}{\hd{Counterfactual}} & \multicolumn{4}{c}{\hd{Ordinary}} &  \\
\cmidrule(lr){2-4}\cmidrule(lr){5-8}
\hd{Method} & \hd{NQ-Swap} & \hd{FaithEval} & \hd{Avg.} & \hd{NQ-Orig.} & \hd{SQuAD} & \hd{HotpotQA} & \hd{Avg.} & \hd{Mean} \\
\midrule
\grouprow{9}{Qwen3-1.7B} \\
Sampling & 62.6 & 58.8 & 60.7 & \textbf{73.3} & 82.0 & \textbf{53.3} & \underline{69.5} & 65.1 \\
CAD & 63.0 & \textbf{63.7} & \textbf{63.3} & 70.6 & 81.3 & 50.1 & 67.3 & 65.3 \\
AdaCAD & \textbf{64.5} & 60.0 & 62.3 & \underline{73.2} & \textbf{82.9} & 52.6 & \textbf{69.6} & \textbf{65.9} \\
Random-passage CD & 62.3 & \underline{63.5} & \underline{62.9} & 71.1 & \underline{82.3} & \underline{52.9} & 68.7 & \underline{65.8} \\
Instructive Decoding & 60.7 & 58.4 & 59.6 & 70.2 & 78.3 & 50.2 & 66.2 & 62.9 \\
\oursrow FCD (ours) & \underline{63.8} & 61.1 & 62.5 & 71.4 & 82.1 & 52.4 & 68.6 & 65.5 \\
\grouprow{9}{Qwen3-4B} \\
Sampling & 60.5 & 63.2 & 61.9 & 74.5 & 85.6 & 63.1 & 74.4 & 68.1 \\
CAD & 63.5 & \textbf{71.6} & 67.6 & 73.6 & 85.8 & 61.3 & 73.6 & \underline{70.6} \\
AdaCAD & 63.2 & 65.0 & 64.1 & 74.6 & 86.4 & 63.0 & 74.7 & 69.4 \\
Random-passage CD & 62.3 & \underline{70.8} & 66.5 & 73.4 & \underline{86.9} & 62.6 & 74.3 & 70.4 \\
Instructive Decoding & \textbf{67.8} & 68.8 & \textbf{68.3} & \textbf{75.8} & 86.0 & \textbf{64.5} & \underline{75.4} & \textbf{71.9} \\
\oursrow FCD (ours) & \underline{65.4} & 70.4 & \underline{67.9} & \underline{75.5} & \textbf{87.6} & \underline{64.3} & \textbf{75.8} & \textbf{71.9} \\
\grouprow{9}{Qwen3-8B} \\
Sampling & 67.7 & 61.5 & 64.6 & \textbf{83.2} & 86.7 & \textbf{69.4} & \textbf{79.8} & 72.2 \\
CAD & 71.4 & \textbf{76.4} & \textbf{73.9} & 81.2 & 86.3 & 67.7 & 78.4 & \textbf{76.2} \\
AdaCAD & 70.3 & 64.0 & 67.1 & \underline{82.9} & 87.2 & \underline{68.4} & \underline{79.5} & 73.3 \\
Random-passage CD & 68.6 & \underline{72.5} & 70.6 & 81.1 & 86.0 & 67.1 & 78.0 & 74.3 \\
Instructive Decoding & \textbf{73.4} & 69.8 & 71.6 & 82.4 & \textbf{88.3} & 67.4 & 79.4 & \underline{75.5} \\
\oursrow FCD (ours) & \underline{72.0} & 71.8 & \underline{71.9} & 82.1 & \underline{87.4} & 67.7 & 79.1 & \underline{75.5} \\
\grouprow{9}{Qwen3-14B} \\
Sampling & 67.9 & 59.0 & 63.4 & 82.5 & 87.3 & \textbf{73.3} & \underline{81.0} & 72.2 \\
CAD & 71.4 & \textbf{73.2} & \textbf{72.3} & 80.4 & 86.4 & 69.3 & 78.7 & \textbf{75.5} \\
AdaCAD & 70.7 & 60.3 & 65.5 & \underline{82.6} & 87.1 & \underline{73.0} & 80.9 & 73.2 \\
Random-passage CD & 68.6 & \underline{70.7} & 69.6 & 80.3 & 86.8 & 71.8 & 79.7 & 74.6 \\
Instructive Decoding & \textbf{74.5} & 63.5 & 69.0 & \textbf{83.5} & \textbf{89.0} & 72.8 & \textbf{81.7} & \underline{75.4} \\
\oursrow FCD (ours) & \underline{72.0} & 67.7 & \underline{69.8} & 82.0 & \underline{87.8} & 72.5 & 80.8 & 75.3 \\
\grouprow{9}{Llama-3.1-8B} \\
Sampling & 46.4 & 72.3 & 59.3 & \textbf{79.4} & 89.1 & \textbf{70.6} & \textbf{79.7} & 69.5 \\
CAD & \underline{58.9} & \textbf{83.2} & \textbf{71.0} & 68.4 & 88.5 & 63.8 & 73.6 & \underline{72.3} \\
AdaCAD & 55.3 & 75.3 & 65.3 & \underline{76.3} & \underline{89.7} & \underline{69.6} & \underline{78.5} & 71.9 \\
Random-passage CD & 54.8 & 77.3 & 66.0 & 75.0 & 88.0 & 66.2 & 76.4 & 71.2 \\
Instructive Decoding & 57.2 & \underline{77.5} & 67.4 & 69.2 & 89.2 & 67.8 & 75.4 & 71.4 \\
\oursrow FCD (ours) & \textbf{62.9} & \underline{77.5} & \underline{70.2} & 75.2 & \textbf{90.0} & 68.7 & 78.0 & \textbf{74.1} \\
\grouprow{9}{Mistral-7B} \\
Sampling & 36.4 & 66.8 & 51.6 & 63.1 & 76.6 & 51.5 & 63.7 & 57.7 \\
CAD & 46.2 & \textbf{74.4} & 60.3 & 60.8 & 79.0 & 52.9 & 64.2 & 62.3 \\
AdaCAD & 44.2 & 68.4 & 56.3 & 63.9 & 80.0 & 52.5 & 65.5 & 60.9 \\
Random-passage CD & 49.6 & \underline{71.9} & \underline{60.8} & \textbf{71.8} & 82.1 & 54.3 & 69.4 & \underline{65.1} \\
Instructive Decoding & \underline{51.9} & 69.0 & 60.4 & 67.6 & \underline{82.7} & \textbf{59.0} & \underline{69.8} & \underline{65.1} \\
\oursrow FCD (ours) & \textbf{53.3} & 71.2 & \textbf{62.2} & \underline{71.4} & \textbf{83.5} & \underline{57.3} & \textbf{70.7} & \textbf{66.5} \\
\bottomrule
}
\caption{Scores of the methods for each model. F1 on free-form tasks, accuracy on FaithEval, and Mean averages the two group averages. Bold marks the best value of a column in a block, underline the second best, and shading our method.}
\label{tab:permodel}
\end{table*}

\section{Guidance Strengths}
\label{app:strengths}

Table~\ref{tab:strengths} lists the views and the strengths chosen for every method on the development split.
FCD tried $\gamma_D\in\{0.25, 0.5, 1, 2, 4, 8\}$ together with $\gamma_E\in\{0.25, 0.5, 1\}$, and Table~\ref{tab:basegrids} lists the strengths that the other methods tried with their scores.

All methods follow one rule.
The score of a strength is the mean of two development gains over sampling, the gain on counterfactual passages and the gain on ordinary passages, each an average over the tasks of its group.
The strength with the highest score is chosen, except that a strength recommended by the original paper is preferred when its score comes within 0.1 points of the highest.
Each method uses one strength for all models and passage tasks.
For AdaCAD the value in the table is an upper bound: at every step the method multiplies it by the Jensen--Shannon divergence between the distributions under the full prompt and under the question alone, which is measured in bits and therefore lies between zero and one.
The alternatives of the ablation follow the same rule: the mean contrast to two other passages tries $\{0.5,1,2\}$, and one view that replaces both factors and one shared strength for both views of FCD try $\{0.25,0.5,1,2\}$.
Paired with the other passage, the directive removed and the instruction removed try $\{0.25,0.5,1,2\}$ while the other passage keeps the strength 0.5, and the question alone tries $\{0.25,0.5,1,2\}$ with $\{0.25,0.5,1\}$ for the other passage.
In the ablation, each view of FCD also appears alone at its strength in FCD, besides the strength that the rule chooses for it as a single-view method.
The strengths of MemoTrap were chosen on its development items in the same way, from $\{0.25,0.5,1,2\}$ for the instruction removed and $\{0.25,0.5,1,2,4\}$ for each of the other two views.
The lower temperature of 0.3 replaces the sampling temperature of the model and uses no guidance.
It is close to the sharpening inside FCD: the score of FCD weights the log-probability of the full prompt by $1+\gamma_D+\gamma_E=\fcdWeight{}$, and the sampling temperatures of Table~\ref{tab:sampling} divided by \fcdWeight{} lie between \lowTDivLo{} and \lowTDivHi{}.
Table~\ref{tab:fcdgrid} lists the development score of every pair of strengths of FCD, which is this mean gain in points over sampling.
The score is flat around the chosen pair: the \gridFlatN{} pairs with a directive strength of \strD{} or twice that lie within \gridFlatGap{} points of the best, so a coarse grid on a small development split is enough to set both strengths.

The effects of the views, measured as in Appendix~\ref{app:additivity}, can also adapt the two strengths of FCD to each model without labels.
Let $A$ and $B$ be the drops in the log-probability of the sampled tokens of a model when the directive and when the evidence are replaced, averaged over the development questions of NQ-Swap, SQuAD and FaithEval, and let $\tilde A$ and $\tilde B$ be their medians over the six models.
The rescaled strengths are $\gamma_D=\strD{}\,\tilde A/A$ and $\gamma_E=\strE{}\,\tilde B/B$, so a factor that moves the tokens of a model less receives a stronger contrast in that model, and Table~\ref{tab:measured} lists them.
They raise the development score from \fcdDev{} to \measDev{} points, and on the test split they add \measGapTwo{} points to the Mean of FCD, with a 95\% interval from \measLoTwo{} to \measHiTwo{}, as Table~\ref{tab:variants} shows.
The same table keeps the strengths of FCD and changes its views: the opposite directive in other words, ``Ignore the passage and answer the question from what you already know'', and the second other passage of the ablation.
Both leave the Mean within \sensMax{} points of FCD, and the 95\% intervals of both differences include zero.

\begin{table*}[!t]
\centering\small
\fulltable{\textwidth}{4}{llcc}{%
\toprule
\hd{Method} & \hd{Contrasted views} & \hd{Strengths} & \hd{Score} \\
\midrule
CAD & question alone & 1 & 2.66 \\
AdaCAD & question alone & 1 & 1.67 \\
Random-passage CD & other passage & 1 & 3.11 \\
Instructive Decoding & opposite directive & 2 & 3.10 \\
\oursrow FCD & opposite directive, other passage & 1, 0.5 & 4.12 \\
\grouprow{4}{Alternatives in the ablation} \\
Two other passages & two other passages & 1 (mean of two contrasts) & 3.34 \\
One view for both factors & opposite directive with other passage & 1 & 3.62 \\
One shared strength & opposite directive, other passage & 1, 1 & 3.84 \\
Directive removed, other passage & directive removed, other passage & 0.5, 0.5 & 3.35 \\
Instruction removed, other passage & instruction removed, other passage & 0.25, 0.5 & 3.60 \\
Question alone, other passage & question alone, other passage & 0.5, 0.5 & 3.52 \\
\grouprow{4}{MemoTrap} \\
Instruction removed & instruction removed & 0.5 & 3.92 \\
Ending of another example & requested ending of another example & 2 & 5.42 \\
Memorized ending & memorized ending & 1 & 6.17 \\
\bottomrule
}
\caption{Views contrasted and guidance strengths chosen on the development split, in the order of the views. The score is the mean of the gain on counterfactual and on ordinary passages over sampling, in points, and for MemoTrap the gain on its development items.}
\label{tab:strengths}
\end{table*}

\begin{table*}[!t]
\centering\small
\fulltable{\textwidth}{3}{lll}{%
\toprule
\hd{Method} & \hd{Strengths tried} & \hd{Scores} \\
\midrule
CAD & 0.25, 0.5, 1, 2 & 1.81, 2.67, \textbf{2.66}, $-$0.49 \\
AdaCAD & 0.5, 1, 2, 4 & 1.26, \textbf{1.67}, 1.54, 0.35 \\
Random-passage CD & 0.25, 0.5, 1, 2 & 1.85, 2.75, \textbf{3.11}, 1.00 \\
Instructive Decoding & 0.25, 0.5, 1, 2, 4, 8 & 0.92, 1.73, 2.82, \textbf{3.10}, 0.76, $-$10.89 \\
\bottomrule
}
\caption{Development score of every strength that a single-view method tried, in the order of the strengths. Bold marks the strength that the rule chose.}
\label{tab:basegrids}
\end{table*}

\begin{table}[t]
\centering\small
\fulltable{\columnwidth}{7}{lcccccc}{%
\toprule
 & \multicolumn{6}{c}{\hd{Directive strength $\gamma_D$}} \\
\cmidrule(lr){2-7}
\hd{$\gamma_E$} & \hd{0.25} & \hd{0.5} & \hd{1} & \hd{2} & \hd{4} & \hd{8} \\
\midrule
0.25 & \cellcolor{heatpos!49}2.71 & \cellcolor{heatpos!57}3.26 & \cellcolor{heatpos!67}3.92 & \cellcolor{heatpos!65}3.80 & \cellcolor{heatpos!35}1.80 & \cellcolor{heatneg!70}$-$8.79 \\
0.5 & \cellcolor{heatpos!58}3.34 & \cellcolor{heatpos!61}3.52 & \cellcolor{heatpos!70}\textbf{4.12} & \cellcolor{heatpos!69}4.07 & \cellcolor{heatpos!41}2.17 & \cellcolor{heatneg!58}$-$7.13 \\
1 & \cellcolor{heatpos!60}3.44 & \cellcolor{heatpos!64}3.71 & \cellcolor{heatpos!66}3.84 & \cellcolor{heatpos!65}3.79 & \cellcolor{heatpos!41}2.22 & \cellcolor{heatneg!48}$-$5.61 \\
\bottomrule
}
\caption{Development score of every pair of strengths of FCD: the mean of the two group gains over sampling, in points. Darker cells mark higher scores, red cells negative ones, and the pair used in the experiments is bold.}
\label{tab:fcdgrid}
\end{table}

\begin{table}[!t]
\centering\small
\fulltable{\columnwidth}{5}{lcccc}{%
\toprule
 & \multicolumn{2}{c}{\hd{Drop (nats)}} & \multicolumn{2}{c}{\hd{Strength}} \\
\cmidrule(lr){2-3}\cmidrule(lr){4-5}
\hd{Model} & \hd{Directive} & \hd{Evidence} & \hd{$\gamma_D$} & \hd{$\gamma_E$} \\
\midrule
Qwen3-1.7B & 0.51 & 4.77 & 0.43 & 0.55 \\
Qwen3-4B & 0.15 & 5.63 & 1.43 & 0.46 \\
Qwen3-8B & 0.29 & 6.03 & 0.74 & 0.43 \\
Qwen3-14B & 0.18 & 5.92 & 1.22 & 0.44 \\
Llama-3.1-8B & 0.26 & 3.22 & 0.85 & 0.81 \\
Mistral-7B & 0.12 & 3.51 & 1.78 & 0.74 \\
\midrule
Median & 0.22 & 5.20 & &  \\
\bottomrule
}
\caption{Mean drop of the log-probability of sampled tokens when the directive or the evidence is replaced, on the development questions of NQ-Swap, SQuAD and FaithEval, and the strengths that rescale those of FCD for each model.}
\label{tab:measured}
\end{table}

\begin{table*}[!t]
\centering\small
\fulltable{\textwidth}{9}{lcccccccc}{%
\toprule
 & \multicolumn{3}{c}{\hd{Counterfactual}} & \multicolumn{4}{c}{\hd{Ordinary}} &  \\
\cmidrule(lr){2-4}\cmidrule(lr){5-8}
\hd{Configuration} & \hd{NQ-Swap} & \hd{FaithEval} & \hd{Avg.} & \hd{NQ-Orig.} & \hd{SQuAD} & \hd{HotpotQA} & \hd{Avg.} & \hd{Mean} \\
\midrule
\oursrow FCD (full method) & 64.9 & 70.0 & 67.4 & 76.3 & 86.4 & 63.8 & 75.5 & 71.5 \\
\grouprow{9}{Views at the strengths of FCD} \\
Opposite directive in other words & 64.5 & 70.3 & 67.4 & 76.6 & 86.3 & 63.3 & 75.4 & 71.4\dfl{$-$0.1} \\
Second other passage & 64.7 & 70.0 & 67.3 & 76.4 & 86.4 & 63.9 & 75.6 & 71.5\dfl{+0.0} \\
\grouprow{9}{Strengths for each model} \\
Rescaled by the measured effects & 65.1 & 70.2 & 67.7 & 76.4 & 86.5 & 63.7 & 75.5 & 71.6\dup{+0.1} \\
\bottomrule
}
\caption{Variants of FCD, averaged over six models. Parentheses give the difference of the Mean to FCD from unrounded scores, colored if its 95\% interval excludes zero, and shading marks FCD.}
\label{tab:variants}
\end{table*}

\section{Effect of the Views}
\label{app:additivity}

To measure how strongly each view changes what the model would say, we follow the trajectories that sampling produces on the development split and record the log-probability of every sampled token under the full prompt and under the views.
Table~\ref{tab:additivity} lists the mean difference to the full prompt when the directive is replaced, when the evidence is replaced and when both are replaced, and the sum of the first two.
The effect of the evidence exceeds that of the directive for every model and task, and the smallest ratio of the two is \addBAMin{}.
Table~\ref{tab:selectivity} compares the same development questions with the swapped passages of NQ-Swap and the original passages of NQ-Original.
From the first to the second, the directive effect falls by \selDirFallMin{} to \selDirFallMax{}\% and the evidence effect by \selEvFallMin{} to \selEvFallMax{}\%, and on every model the directive effect falls by the larger share.
For each model the fall of the directive effect has a 95\% interval above zero, computed from 10,000 bootstrap resamples of the questions.
The ordinary passages of SQuAD, which come with other questions, show the same pattern: the median directive effect is \addDirDrop{}\% lower there than on NQ-Swap, at \addDirSq{} nats, whereas the median evidence effect is \addEvDrop{}\% lower, at \addEvSq{} nats.
The directive view thus acts mainly where the passage contradicts what the model memorized.
Replacing both factors lowers the log-probability by \addRatio{} times the sum of the two single effects, as a median over models and tasks, and the two follow each other with a median correlation of \addR{} over examples.
The joint drop falls short of the sum because the opposite directive, which tells the model to ignore the passage, has little left to change once the passage carries no information about the question.

These averages run over every token of an answer, including tokens that follow from the format or from the words already written.
To look at the decision itself, we score two answers to each NQ-Swap question of the development split, the answer that the passage supports and the memorized answer, by the sum of their token log-probabilities under the full prompt and under the views, and take the log-odds of the first against the second.
Table~\ref{tab:decision} lists the median over questions of the drop of this log-odds when a factor is replaced.
The evidence replacement lowers it by \dmEvMin{} to \dmEvMax{} nats and the directive replacement by \dmDirMin{} to \dmDirMax{} nats, so the evidence effect is the larger one for every model, by a median factor of \dmRatio{}.
The question alone lowers the log-odds by \dmQMin{} to \dmQMax{} nats, and the evidence replacement alone accounts for a median of \dmEvShare{}\% of that drop.

Sampled from the question alone, the view used by CAD, the answers to the NQ-Swap test questions reach an F1 of \soloOrg{} against the original answer and \soloSub{} against the substituted one, averaged over the six models, and the score against the original answer is higher for \soloPosModels{}.
A question counts as known to a model when these answers reach a mean F1 of at least 0.5 against the original answer over the three seeds, and for such a question the passage contradicts what the model memorized.
On the known questions FCD gains \soloGainKnown{} points over sampling and on the other questions \soloGainOther{} points, with a 95\% interval of the difference from \soloGainLo{} to \soloGainHi{}.

\begin{table}[!t]
\centering\small
\fulltable{\columnwidth}{6}{lccccc}{%
\toprule
\hd{Model} & \hd{Directive} & \hd{Evidence} & \hd{Both} & \hd{Sum} & \hd{$r$} \\
\midrule
\grouprow{6}{NQ-Swap} \\
Qwen3-1.7B & 0.39 & 6.76 & 6.63 & 7.16 & 0.92 \\
Qwen3-4B & 0.35 & 8.30 & 6.28 & 8.65 & 0.89 \\
Qwen3-8B & 0.52 & 7.88 & 6.42 & 8.40 & 0.92 \\
Qwen3-14B & 0.31 & 7.72 & 6.26 & 8.02 & 0.93 \\
Llama-3.1-8B & 0.59 & 4.21 & 3.28 & 4.80 & 0.79 \\
Mistral-7B & 0.26 & 4.71 & 4.39 & 4.98 & 0.96 \\
\grouprow{6}{SQuAD} \\
Qwen3-1.7B & 0.55 & 5.75 & 6.32 & 6.30 & 0.90 \\
Qwen3-4B & 0.08 & 6.94 & 5.11 & 7.03 & 0.81 \\
Qwen3-8B & 0.23 & 6.31 & 4.89 & 6.54 & 0.86 \\
Qwen3-14B & 0.08 & 5.73 & 4.53 & 5.80 & 0.86 \\
Llama-3.1-8B & 0.07 & 3.56 & 2.01 & 3.63 & 0.77 \\
Mistral-7B & 0.09 & 4.86 & 4.37 & 4.95 & 0.94 \\
\bottomrule
}
\caption{Drop in the log-probability of the sampled tokens when the directive, the evidence or both are replaced, the sum of the first two, and the correlation $r$ between the drop for both and that sum over examples.}
\label{tab:additivity}
\end{table}

\begin{table}[!t]
\centering\small
\fulltable{\columnwidth}{7}{lcccccc}{%
\toprule
 & \multicolumn{3}{c}{\hd{Directive}} & \multicolumn{3}{c}{\hd{Evidence}} \\
\cmidrule(lr){2-4}\cmidrule(lr){5-7}
\hd{Model} & \hd{Swap} & \hd{Orig.} & \hd{Fall} & \hd{Swap} & \hd{Orig.} & \hd{Fall} \\
\midrule
Qwen3-1.7B & 0.39 & 0.16 & 59\% & 6.76 & 5.37 & 21\% \\
Qwen3-4B & 0.35 & 0.15 & 56\% & 8.30 & 6.71 & 19\% \\
Qwen3-8B & 0.52 & 0.19 & 65\% & 7.88 & 5.44 & 31\% \\
Qwen3-14B & 0.31 & 0.12 & 60\% & 7.72 & 4.56 & 41\% \\
Llama-3.1-8B & 0.59 & 0.11 & 82\% & 4.21 & 2.94 & 30\% \\
Mistral-7B & 0.26 & 0.15 & 41\% & 4.71 & 4.37 & 7\% \\
\bottomrule
}
\caption{Drop in the log-probability of the sampled tokens when the directive or the evidence is replaced, for the same development questions with the swapped passage of NQ-Swap and the original passage of NQ-Original, and the share by which each drop falls.}
\label{tab:selectivity}
\end{table}

\begin{table*}[!t]
\centering\small
\fulltable{\textwidth}{7}{lcccccc}{%
\toprule
\hd{Model} & \hd{Full prompt} & \hd{Directive replaced} & \hd{Evidence replaced} & \hd{Both replaced} & \hd{Question alone} & \hd{Ratio} \\
\midrule
Qwen3-1.7B & 28.8 & 1.4 & 35.1 & 37.3 & 41.0 & 24.7 \\
Qwen3-4B & 28.5 & 4.2 & 36.2 & 38.3 & 42.7 & 8.7 \\
Qwen3-8B & 29.0 & 8.4 & 41.3 & 45.1 & 48.4 & 4.9 \\
Qwen3-14B & 23.7 & 4.7 & 39.6 & 41.2 & 42.8 & 8.4 \\
Llama-3.1-8B & 6.2 & 5.8 & 16.5 & 22.4 & 23.5 & 2.9 \\
Mistral-7B & 13.3 & 3.9 & 24.7 & 28.6 & 35.7 & 6.3 \\
\midrule
Median & 26.1 & 4.5 & 35.7 & 37.8 & 41.9 & 7.3 \\
\bottomrule
}
\caption{Log-odds of the passage answer against the memorized answer for NQ-Swap questions of the development split under the full prompt, and the drop caused by each replacement, in nats. Ratio divides the drop for the evidence by the drop for the directive.}
\label{tab:decision}
\end{table*}

\begin{table*}[!t]
\centering\small
\fulltable{\textwidth}{6}{lccccc}{%
\toprule
 & \multicolumn{3}{c}{\hd{Exact match}} & \multicolumn{2}{c}{\hd{Tokens}} \\
\cmidrule(lr){2-4}\cmidrule(lr){5-6}
\hd{Method} & \hd{Counterfactual} & \hd{Ordinary} & \hd{Mean} & \hd{Mistral-7B} & \hd{Other five} \\
\midrule
Sampling & 56.6 & 61.3 & 59.0 & 10.4 & 4.8 \\
CAD & \underline{63.2} & 58.5 & 60.9 & 9.9 & 5.2 \\
AdaCAD & 59.7 & 61.7 & 60.7 & 9.6 & 4.8 \\
Random-passage CD & 61.1 & 60.0 & 60.6 & 8.4 & 5.1 \\
Instructive Decoding & 62.4 & \underline{62.0} & \underline{62.2} & 7.2 & 4.5 \\
\oursrow FCD (ours) & \textbf{63.3} & \textbf{62.4} & \textbf{62.8} & 7.5 & 4.7 \\
\bottomrule
}
\caption{Exact match, and the mean answer length on ordinary passages in tokens. Bold marks the best exact-match value of a column, underline the second best, and shading FCD.}
\label{tab:emlength}
\end{table*}

\begin{table}[!t]
\centering\small
\fulltable{\columnwidth}{5}{lcccc}{%
\toprule
 & \multicolumn{2}{c}{\hd{Counterfactual}} & \multicolumn{2}{c}{\hd{Ordinary}} \\
\cmidrule(lr){2-3}\cmidrule(lr){4-5}
\hd{Method} & \hd{Repaired} & \hd{Broken} & \hd{Repaired} & \hd{Broken} \\
\midrule
CAD & 26.0 & 3.1 & 15.0 & 7.1 \\
AdaCAD & 10.5 & 1.6 & 9.3 & 2.8 \\
Random-passage CD & 20.8 & 3.0 & 17.5 & 5.0 \\
Instructive Decoding & 21.5 & 3.7 & 20.1 & 5.3 \\
\oursrow FCD (ours) & 22.1 & 1.8 & 17.9 & 3.7 \\
\bottomrule
}
\caption{Shares of answers that a method repairs or breaks relative to sampling, in percent of the answers that sampling gets wrong or right. Shading marks FCD.}
\label{tab:flips}
\end{table}

\section{Exact Match, Answer Length and Changed Answers}
\label{app:em}

Table~\ref{tab:emlength} lists exact match, which credits an answer only when it equals a reference answer after normalization, and the mean length of the answers to ordinary passages.
Under exact match the 95\% interval of the difference between FCD and Instructive Decoding, the method closest to it, runs from \emGapIdLo{} to \emGapIdHi{} points on the Mean, and that between FCD and CAD from \emGapCadLo{} to \emGapCadHi{} points.
Mistral-7B answers with \tokMistralBase{} tokens on average under sampling, against \tokOtherBase{} for the other five models.
Every guidance method shortens the answers of Mistral-7B, which raises both scores on ordinary passages, and FCD shortens them to \tokMistralOurs{} tokens while it leaves the length of the answers of the other five models at \tokOtherOurs{} tokens.
Without Mistral-7B, FCD remains ahead of every other method on the Mean; its smallest lead is \emFiveFMin{} points under F1, with intervals above zero, and \emFiveEmMin{} points under exact match.

Table~\ref{tab:flips} compares the answers of every method with those of sampling, pooled over the six models and three seeds.
An answer is right when its F1 reaches \rightThr{} or, for FaithEval, when the chosen letter is correct.
A method repairs an answer that sampling gets wrong and breaks an answer that sampling gets right, and the shares are taken over the wrong and the right answers of sampling.
On counterfactual passages FCD repairs \repOursCf{}\% of the wrong answers and breaks \brkOursCf{}\% of the right ones, and on ordinary passages it repairs \repOursOr{}\% and breaks \brkOursOr{}\%, against \repCadOr{}\% and \brkCadOr{}\% for CAD.

\section{Additional Baselines and Settings}
\label{app:extra}
\begin{table*}[!t]
\centering\small
\fulltable{\textwidth}{9}{lcccccccc}{%
\toprule
 & \multicolumn{3}{c}{\hd{Counterfactual}} & \multicolumn{4}{c}{\hd{Ordinary}} &  \\
\cmidrule(lr){2-4}\cmidrule(lr){5-8}
\hd{Method} & \hd{NQ-Swap} & \hd{FaithEval} & \hd{Avg.} & \hd{NQ-Orig.} & \hd{SQuAD} & \hd{HotpotQA} & \hd{Avg.} & \hd{Mean} \\
\midrule
\grouprow{9}{Mean over six models} \\
Sampling & 56.9 & 63.6 & 60.3 & \underline{76.0} & 84.5 & 63.5 & 74.7 & 67.5 \\
CAD & 62.4 & \textbf{73.7} & \textbf{68.1} & 72.5 & 84.6 & 60.8 & 72.6 & 70.3 \\
AdaCAD & 61.4 & 65.5 & 63.4 & 75.6 & 85.5 & 63.2 & \underline{74.8} & 69.1 \\
Random-passage CD & 61.0 & \underline{71.1} & 66.1 & 75.4 & 85.3 & 62.5 & 74.4 & 70.2 \\
Instructive Decoding & \underline{64.3} & 67.8 & 66.0 & 74.8 & \underline{85.6} & 63.6 & 74.7 & \underline{70.4} \\
NWCAD & 57.3 & 63.8 & 60.6 & \underline{76.0} & 84.5 & \underline{63.7} & \underline{74.8} & 67.7 \\
COIECD & 62.2 & \textbf{73.7} & \underline{68.0} & 72.7 & 84.5 & 61.1 & 72.8 & \underline{70.4} \\
\oursrow FCD (ours) & \textbf{64.9} & 70.0 & 67.4 & \textbf{76.3} & \textbf{86.4} & \textbf{63.8} & \textbf{75.5} & \textbf{71.5} \\
\grouprow{9}{Qwen3-32B} \\
Sampling & 65.8 & 58.7 & 62.2 & \textbf{83.7} & 89.4 & \textbf{73.8} & \textbf{82.3} & 72.3 \\
CAD & \underline{69.5} & \textbf{74.5} & \textbf{72.0} & 77.2 & 88.4 & 64.4 & 76.7 & \underline{74.4} \\
AdaCAD & \textbf{71.3} & 61.5 & 66.4 & \underline{83.4} & \underline{90.0} & 71.6 & \underline{81.6} & 74.0 \\
Random-passage CD & 66.3 & 71.0 & 68.6 & 78.0 & 89.2 & 68.1 & 78.4 & 73.5 \\
Instructive Decoding & 65.9 & 67.8 & 66.8 & 75.3 & 88.5 & \underline{72.3} & 78.7 & 72.8 \\
\oursrow FCD (ours) & 69.1 & \underline{71.4} & \underline{70.3} & 79.4 & \textbf{90.2} & 70.7 & 80.1 & \textbf{75.2} \\
\bottomrule
}
\caption{Six-model average with NWCAD and COIECD added, and Qwen3-32B. F1 on free-form tasks, accuracy on FaithEval; Mean averages the two groups. In each block, bold marks the best column value, underline the second best, and shading our method.}
\label{tab:extra}
\end{table*}

\begin{table}[!t]
\centering\small
\fulltable{\columnwidth}{4}{lccc}{%
\toprule
\hd{Setting} & \hd{Counterfactual} & \hd{Ordinary} & \hd{Score} \\
\midrule
\grouprow{4}{NWCAD, divergence threshold} \\
0.05 & 0.33 & \textbf{0.02} & 0.17 \\
\rowcolor{oursbg}0.1 & \textbf{0.46} & 0.00 & \textbf{0.23} \\
0.2 & 0.44 & $-$0.05 & 0.20 \\
0.3 & 0.33 & $-$0.65 & $-$0.16 \\
0.5 & $-$0.26 & $-$1.26 & $-$0.76 \\
\grouprow{4}{COIECD, strength and constraint scale} \\
0.5, 0.25 & $-$21.00 & $-$4.09 & $-$12.55 \\
1, 0.25 & 2.01 & 0.36 & 1.19 \\
\rowcolor{oursbg}2, 0.25 & \textbf{7.43} & $-$2.23 & \textbf{2.60} \\
1, 0.1 & 2.70 & \textbf{0.37} & 1.54 \\
1, 0.5 & 0.99 & 0.04 & 0.51 \\
\bottomrule
}
\caption{Development scores of the settings tried for NWCAD and COIECD: the gain over sampling in points on counterfactual and on ordinary passages and the mean of the two. The best value of a block is bold and the shaded row is the setting used.}
\label{tab:extragrid}
\end{table}

\begin{table*}[!t]
\centering\small
\fulltable{\textwidth}{9}{lcccccccc}{%
\toprule
 & \multicolumn{3}{c}{\hd{Counterfactual}} & \multicolumn{4}{c}{\hd{Ordinary}} &  \\
\cmidrule(lr){2-4}\cmidrule(lr){5-8}
\hd{Method} & \hd{NQ-Swap} & \hd{FaithEval} & \hd{Avg.} & \hd{NQ-Orig.} & \hd{SQuAD} & \hd{HotpotQA} & \hd{Avg.} & \hd{Mean} \\
\midrule
Greedy decoding & 42.4 & 69.5 & 55.9 & \underline{71.7} & 83.1 & 62.1 & 72.3 & 64.1 \\
CAD & 52.8 & \textbf{78.8} & \underline{65.8}\dup{+9.9} & 64.7 & 83.5 & 58.9 & 69.0\ddn{$-$3.3} & 67.4\dup{+3.3} \\
AdaCAD & 50.1 & 71.8 & 61.0\dup{+5.0} & 70.4 & 85.1 & 61.6 & 72.4\dfl{+0.1} & 66.7\dup{+2.6} \\
Random-passage CD & 52.9 & \underline{75.0} & 64.0\dup{+8.0} & \textbf{73.9} & 85.2 & 60.3 & 73.2\dup{+0.8} & \underline{68.6}\dup{+4.4} \\
Instructive Decoding & \underline{54.5} & 73.2 & 63.9\dup{+7.9} & 69.4 & \underline{86.6} & \textbf{64.0} & \underline{73.4}\dup{+1.0} & \underline{68.6}\dup{+4.5} \\
\oursrow FCD (ours) & \textbf{59.0} & 74.1 & \textbf{66.6}\dup{+10.6} & \textbf{73.9} & \textbf{86.7} & \underline{63.3} & \textbf{74.6}\dup{+2.3} & \textbf{70.6}\dup{+6.5} \\
\bottomrule
}
\caption{Scores under greedy decoding, averaged over Llama-3.1-8B and Mistral-7B, with the strengths of the main experiments. Parentheses give the difference to unguided greedy decoding, colored if its 95\% interval excludes zero. Bold marks the best value, underline the second best.}
\label{tab:greedy}
\end{table*}

\begin{table*}[!t]
\centering\small
\fulltable{\textwidth}{8}{lccccccc}{%
\toprule
\hd{Method} & \hd{Qwen3-1.7B} & \hd{Qwen3-4B} & \hd{Qwen3-8B} & \hd{Qwen3-14B} & \hd{Llama-3.1-8B} & \hd{Mistral-7B} & \hd{Avg.} \\
\midrule
Sampling & 44.2 & 45.9 & 47.5 & 44.0 & 43.1 & 40.1 & 44.1 \\
\oursrow FCD (ours) & 50.3 & 51.4 & 53.2 & 47.2 & 54.8 & 43.0 & 50.0\dup{+5.9} \\
\bottomrule
}
\caption{Accuracy on the FaithEval passages that lack the answer, where the right reply states that the answer is missing. Parentheses give the difference to sampling, colored if its 95\% interval excludes zero, and shading marks our method.}
\label{tab:unans}
\end{table*}

\begin{table*}[!t]
\centering\small
\fulltable{\textwidth}{11}{lcccccccccc}{%
\toprule
 & \multicolumn{4}{c}{\hd{Mean over models}} & \multicolumn{6}{c}{\hd{Average of the subsets}} \\
\cmidrule(lr){2-5}\cmidrule(lr){6-11}
\hd{Method} & \hd{QA} & \hd{MR} & \hd{MC} & \hd{Avg.} & \hd{Q-1.7B} & \hd{Q-4B} & \hd{Q-8B} & \hd{Q-14B} & \hd{Llama} & \hd{Mistral} \\
\midrule
Sampling & 71.6 & 61.9 & 55.9 & 63.1 & 72.0 & 71.1 & 71.7 & 69.5 & 54.2 & 40.4 \\
CAD & 75.6 & 65.6 & 58.6 & 66.6 & 73.0 & 74.2 & 74.2 & 72.4 & 54.1 & 51.8 \\
AdaCAD & 73.9 & 64.3 & 58.7 & 65.7 & 72.9 & 72.9 & 73.4 & 71.8 & 55.7 & 47.2 \\
Random-passage CD & 75.9 & 66.2 & 61.2 & 67.8 & 72.3 & 73.9 & 73.3 & 72.7 & \underline{60.9} & 53.4 \\
Instructive Decoding & \underline{77.4} & \underline{67.8} & \underline{63.1} & \underline{69.4} & \underline{73.1} & \underline{75.1} & \textbf{77.1} & \textbf{76.4} & 58.8 & \underline{56.2} \\
\oursrow FCD (ours) & \textbf{78.1} & \textbf{68.5} & \textbf{64.2} & \textbf{70.3} & \textbf{73.4} & \textbf{76.0} & \underline{75.8} & \underline{74.8} & \textbf{65.2} & \textbf{56.4} \\
\bottomrule
}
\caption{Held-out ConFiQA subsets, test split, F1 against the passage answer: single facts (QA), facts on a chain of reasoning (MR) and several changed facts (MC). Q- abbreviates Qwen3. Bold marks the best column value, underline the second best.}
\label{tab:heldout}
\end{table*}

We add two recent methods that decide at every step how to apply the contrast of CAD, a larger model, greedy decoding and passages that lack the answer.
No-Worse CAD (NWCAD) \citep{tao2026noworse} contrasts with the question alone, as CAD does.
NWCAD decodes from the question-alone distribution if it is confident and stays near the full-prompt distribution, from the full-prompt distribution if that one is confident, and otherwise from the CAD score.
COIECD \citep{yuan-etal-2024-discerning} checks every token against an entropy constraint on the full-prompt distribution: a token that breaks the constraint gets the CAD score, and every other token gets its log-probability under the question alone plus the contrast.
Our implementations follow the two papers, and for COIECD also the decoding code that its authors released.
For NWCAD we keep the published confidence margins, 0.30 for the question alone and 0.05 for the full prompt, and compute the divergence over the whole vocabulary.

Table~\ref{tab:extra} shows the six main models in its upper block, with the settings that the development split selects for the two methods.
NWCAD stays close to sampling and changes counterfactual passages by \nwdCf{} points and ordinary passages by \nwdOr{} points.
COIECD changes them by \codCf{} and \codOr{} points, against \dCadCf{} and \dCadOr{} points for CAD.
FCD is ahead of NWCAD and COIECD on the Mean by \nwGapBoth{} and \coGapBoth{} points, with 95\% intervals from \nwVsBothLo{} to \nwVsBothHi{} and from \coVsBothLo{} to \coVsBothHi{}.
The lower block shows Qwen3-32B in non-thinking mode, which has more than twice the parameters of the largest of the six models, with the methods and the settings of the main experiments, and FCD has the highest Mean there as well.

The development scores of every setting that we tried for the two methods are in Table~\ref{tab:extragrid}, computed with the rule of Appendix~\ref{app:strengths}.
For NWCAD the rule selects a divergence threshold of \nwTau{}, and for COIECD a strength of \coAlpha{} with a constraint scale of \coLam{}.
The settings that the original papers recommend, a threshold of 0.3 for NWCAD and a strength of 1 with a constraint scale of 0.25 for COIECD, score \xgNwGap{} and \xgCoGap{} points lower.

Table~\ref{tab:greedy} repeats the comparison of Table~\ref{tab:main} on Llama-3.1-8B and Mistral-7B with greedy decoding, where every step emits the highest-scoring token.
The Qwen3 models are left out because their model card advises against greedy decoding.
Every method keeps the strengths of the main experiments, and each setting runs once because greedy decoding involves no randomness.
FCD has the highest Mean on \grdTopModels{}, and on their average it gains \grdOursBoth{} points over greedy decoding without guidance and leads every other method by \grdGapMin{} to \grdGapMax{} points, with 95\% intervals above zero.

FaithEval also offers passages that lack the answer to the question \citep{ming2025faitheval}.
Their instruction adds a request to reply unknown when the passage does not contain the answer, and an answer is right when its first line contains, as whole words, one of the keywords that the dataset lists for such answers, such as unknown or no answer.
The pool and the splits are drawn as for the other tasks, and FCD keeps its views and strengths without any tuning on these passages.
Table~\ref{tab:unans} shows that FCD raises this accuracy from \unSamp{} to \unOurs{}\%, a gain of \unGain{} points with a 95\% interval from \unGainLo{} to \unGainHi{}, and that it is above sampling on \unPosModels{}.

ConFiQA \citep{bi2025context} builds counterfactual passages from Wikidata facts: its QA subset changes the fact that answers a single-hop question, and its MR and MC subsets change one fact or several facts on the chain behind a multi-hop question.
Two filters apply: the changed answer and the original answer must share no alias, and the passage may have at most 1,000 words.
From what remains of the test set of each subset we draw the pool and the splits as for the other tasks.
F1 is computed against the answer supported by the changed passage, and every strength stays as in the main experiments.
Table~\ref{tab:heldout} lists the three subsets averaged over the six models and the average of the subsets for each model.
FCD has the highest average, \hoGainBase{} points above sampling and \hoGapBest{} points above \hoBest{} with a 95\% interval from \hoGapLo{} to \hoGapHi{}, and it has the highest average of the subsets on \hoTopModels{}.

\end{document}